\documentclass[accepted]{uai2025}
                        
\usepackage[american]{babel}
\usepackage{amssymb}
\usepackage{graphicx}
\usepackage{epsfig}
\usepackage{epstopdf}
\usepackage[round]{natbib}
\usepackage{mathtools} 
\usepackage{booktabs} 
\usepackage{tikz} 
\DeclareMathOperator*{\argmax}{argmax}
\usepackage{algorithm, bbm}
\usepackage{algorithmic}
\usepackage{subcaption}

\newtheorem{theorem}{Theorem}[section]
\newtheorem{lemma}{Lemma}[section]
\newtheorem{proposition}{Proposition}[section]

\newtheorem{remark}{Remark}[section]
\numberwithin{equation}{section}

\newtheorem{assumption}{Assumption}[section]

\title{Best Arm Identification with Possibly Biased Offline Data}

\author[1]{\href{mailto:<le_yang@nus.edu.sg>?Subject=Your UAI 2025 paper}{Le Yang}}
\author[3,1]{Vincent~Y.~F.~Tan}
\author[2]{Wang Chi Cheung}
\affil[1]{%
    Department of Electrical and Computer Engineering\\
    National University of Singapore\\
    Singapore}
\affil[2]{%
    Department of Industrial Systems Engineering and Management\\
    National University of Singapore\\
    Singapore}
\affil[3]{%
    Department of Mathematics\\
    National University of Singapore\\
    Singapore}  
    
\begin{document}
\maketitle

\begin{abstract}
We study the best arm identification (BAI) problem with potentially biased offline data in the fixed confidence setting, which commonly arises in real-world scenarios such as clinical trials. We prove an impossibility result for adaptive algorithms without prior knowledge of the bias bound between online and offline distributions. To address this, we propose the LUCB-H algorithm, which introduces adaptive confidence bounds by incorporating an auxiliary bias correction to balance offline and online data within the LUCB framework. Theoretical analysis shows that LUCB-H matches the sample complexity of standard LUCB when offline data is misleading and significantly outperforms it when offline data is helpful. We also derive an instance-dependent lower bound that matches the upper bound of LUCB-H in certain scenarios. Numerical experiments further demonstrate the robustness and adaptability of LUCB-H in effectively incorporating offline data.
\end{abstract}

\section{Introduction}
Best Arm Identification (BAI) is a fundamental problem in multi-armed bandit (MAB) research, where the goal is to find the arm with the highest expected reward through sequential sampling. Traditionally, BAI algorithms rely only on data collected during online. By contrast, in many practical applications, the decision maker (DM) has access to substantial amounts of historical data. Historical data, if relevant, can significantly reduce the number of online samples needed for decision-making if used correctly. Unfortunately, historical data often comes from different environments, making them biased and potentially misleading.

Challenges like this arise frequently in real-world applications. For example, in personalized recommendation systems, companies often possess years of historical user interaction data, which could provide a strong starting point for new recommendation models. However, user preferences and behavior evolve over time, creating a gap between past and present data distributions. Similarly, in clinical trials, previous studies can offer useful insights into treatment effectiveness, but differences in patient demographics or clinical settings may introduce biases that could mislead decision-making.

These examples highlight the importance of developing robust algorithms that can adaptively decide when and how to incorporate historical data. However, no existing algorithm is designed for BAI with possibly biased offline data. In this paper, we bridge this gap by proposing a novel approach.

Inspired by the above discussion, we consider BAI with possibly biased offline data in a fixed confidence setting. The learning process consists of two phases: a warm-start phase and an online phase. In the warm-start phase, the DM receives an offline dataset generated by a latent distribution $P^{\text{off}}$, which can be used to accelerate the identification process in the subsequent online phase. In the online phase, rewards are drawn from another latent distribution $P^{\text{on}}$. The goal is to identify the arm with the highest mean reward at a fixed confidence level while minimizing the sample complexity, specifically the number of samples collected during the online phase.

Intuitively, when $P^{\mathrm{off}}$ and $P^{\mathrm{on}}$ are ``far apart", the DM should ignore the offline dataset and conduct online learning from scratch. For example, the Track-and-Stop policy \citep{garivier2016optimal} and the LUCB algorithm \citep{kalyanakrishnan2012pac} require  
\begin{align}\label{lower1}
    \Theta\left(\sum_{i:\Delta_{i}>0}\Delta_{i}^{-2}\log\left(\frac{1}{\delta}\right)\right)
\end{align}  
samples to ensure correct identification with probability at least $1-\delta$, where $\Delta_{i} > 0$ is the suboptimality gap between the expected reward of the optimal arm and arm $i$ and $\delta \in (0, 1)$ is the given confidence level. This lower bound holds for any $P^{\mathrm{off}}$ and $P^{\mathrm{on}}$.  

In contrast, when $P^{\mathrm{off}}$ and $P^{\mathrm{on}}$ are ``sufficiently close'' the DM can incorporate offline data to reduce unnecessary exploration. For instance, when $P^{\mathrm{off}} = P^{\mathrm{on}}$, the batch Track-and-Stop policy \citep{agrawal2023optimal} requires  
\begin{align}\label{lower2}    \Theta\left(\sum_{i:\Delta_{i}>0}\max\left(\Delta_{i}^{-2}\log\left(\frac{1}{\delta}\right) - T_{S}(i), 0\right)\right)
\end{align}  
samples to achieve the same level of confidence, where $T_{S}(i)$ represents the number of offline samples for arm $i$ in the warm-start phase. This bound is tighter than (\ref{lower1}), but it only applies when $P^{\mathrm{off}} = P^{\mathrm{on}}$.



The goal of this paper is to design a BAI algorithm that adaptively decides whether to utilize historical data in each iteration, achieving a tighter bound than (\ref{lower1}) when $P^{\mathrm{off}} = P^{\mathrm{on}}$, while retaining the bound (\ref{lower1}) for general (not necessarily equal) $P^{\mathrm{off}}$ and~$P^{\mathrm{on}}$.

In this paper, we first reveal the fundamental limitations of adaptive algorithms from a theoretical perspective. Specifically, we construct two instances to demonstrate an impossibility result: without prior knowledge of an upper bound on the bias between offline and online data, any adaptive algorithm leveraging offline data may suffer higher sample complexity in certain cases, even exceeding the optimal sample complexity of purely online strategies. This result shows that it is infeasible to fully adaptively decide when and how to use offline data with unknown bias, highlighting the need for additional auxiliary information to design efficient algorithms.

To address this problem, we propose the Lower and Upper Confidence Bounds - History (LUCB-H) algorithm. LUCB-H introduces an effective bias upper bound to achieve adaptive correction of offline data within the LUCB framework. In each iteration, it computes upper and lower confidence bounds based on both online data and a combination of online and offline data. The algorithm then selects the smaller upper confidence bound and the larger lower confidence bound to obtain more accurate estimates, dynamically deciding whether to incorporate offline data. The algorithm significantly reduces sample complexity when offline data is helpful while discarding the data when the information is misleading, ensuring that the sample complexity remains consistent with that of the standard LUCB algorithm.  

We present an instance-dependent lower bound on the sample complexity of $\delta$-PAC algorithm, which can significantly reduce the sample complexity when the offline data is helpful for identifying the best arm. Moreover, we prove that LUCB-H can achieve this lower bound in certain specific instances, indicating that the algorithm is theoretically near-optimal in terms of sample complexity. Numerical experiment results further validate the effectiveness and robustness of LUCB-H, especially in complex environments with potentially biased offline data.  

\subsection{Related Work}
For the MAB problem, much attention has been devoted to the integration of offline data into online learning to improve performance guarantees. The majority of existing studies focus on the specific setting where offline and online share the same distribution ($P^{\mathrm{off}} = P^{\mathrm{on}}$) ensuring that historical data can be safely leveraged.
\citet{shivaswamy2012multi} integrated historical data in MAB, showing that logarithmic historical data reduces regret from logarithmic to constant. \citet{agrawal2023optimal} extended Track-and-Stop \citep{garivier2016optimal}, achieving asymptotical optimal sample complexity under aligned offline and online distributions for BAI. \citet{liu2025offline} focused on the integration of historical data in combinatorial bandits, and demonstrated its effectiveness in practical applications. Other research has focused on leveraging offline data in domains such as dynamic pricing \citep{bu2020online}, clustered bandits \citep{bouneffouf2019optimal, ye2020combining}, reinforcement learning \citep{hao2023leveraging, wagenmaker2023leveraging}, and sequential data settings \citep{gur2022adaptive}.

Thompson Sampling (TS) is a popular Bayesian method that effectively incorporates offline data into online learning by constructing prior distributions, yielding significant performance gains when distributions align \citep{russo2016information}. However, prior mis-specification can severely affect performance, making TS worse than state-of-the-art online policies without any offline data \citep{liu2016prior, simchowitz2021bayesian}. Such results illustrate the necessity of adaptive strategies for deciding whether and how historical data should be used, particularly in scenarios with potentially biased offline data.

\citet{zhang2019warm} explored the integration of biased offline data into contextual bandits. They proposed the ARROW-CB algorithm, which uses a weighted strategy to balance between ignoring and fully incorporating offline data without requiring prior knowledge of the offline-online discrepancy. While ARROW-CB does not offer improved regret bounds compared to baseline algorithms that ignore offline data. \citet{cheung2024leveraging} formally demonstrated that, in the absence of a non-trivial upper bound on the discrepancy, no non-anticipatory policy can achieve better regret bounds than such baselines. To address this issue, \citet{cheung2024leveraging} introduced an online policy MIN-UCB, which incorporates a non-trivial bias bound that adaptively determines whether to incorporate or discard historical data, improving regret bounds when historical data is helpful and retaining regret bounds when it is misleading. In contrast to our work, MIN-UCB focuses on regret minimization in MAB, while we address the BAI problem.

The discrepancy between offline and online data distributions plays a crucial role in determining whether historical data can be effectively leveraged. Several studies have explored strategies to address this challenge. \citet{si2023distributionally} proposed a distributionally robust strategy for learning under environmental changes, while \citet{chen2022data} explored distribution shift in reinforcement learning. Similar studies exist in supervised learning, such as \citet{crammer2008learning,mansour2009domain,ben2010theory}.


\section{Formulation}
We consider the BAI problem with possibly biased offline data. Let $A = \{1, 2, \dots, k\}$ be the set of arms. The learning process consists of two phases: the warm-start phase and the online phase. During the warm-start phase, DM receives historical reward data for each arm. Specifically, for each arm $i \in A$, the DM is given $T_S(i)$ independent and identically distributed (i.i.d.) samples denoted as $\{X_s(i)\}_{s=1}^{T_S(i)}$, where each sample is drawn from an offline reward distribution $P_i^{\mathrm{off}}$.

In the subsequent online phase, the DM selects an arm $A_t \in A$ to pull at each round $t$, and observes a stochastic reward $Y_t(A_t)$ drawn from the online reward distribution $P_{A_t}^{\mathrm{on}}$. The offline and online reward distributions may differ, i.e., $P^{\mathrm{off}} \neq P^{\mathrm{on}}$. 

The DM follows a policy $\pi = (\pi_t)_{t=1}^{\infty}$, which is a (possibly randomized) rule for selecting an arm $A_t$ at each round $t$. The decision is based on the available observations up to time $t-1$, given by $\mathcal{F}_{t-1}=(S=\{T_{S}(i)\}_{i\in A}, \{A_l, Y_l(A_l)\}_{l=1}^{t-1})$. 
The underlying instance $I$ is defined as a four-tuple $(A, \{T_{S}(i)\}_{i\in A}, P, \delta)$, where $A$ is the set of arms, $\{T_{S}(i)\}_{i\in A}$ represents the number of historical reward data collected from the warm-start phase for each arm $i$, $P$ denotes the reward distribution consisting of the online and offline reward distributions $P^{\mathrm{on}}$ and $P^{\mathrm{off}}$, and $\delta$ is a given constant. In this setup, the DM has access only to $A$ and $\{T_{S}(i)\}_{i\in A}$ before the online phase, whereas $P$ and $\delta$ are unknown. In this paper, we assume that both the online and offline reward distributions, $P^{\mathrm{on}}$ and $P^{\mathrm{off}}$, are normal distributions with known variance, a fact that is also known to the DM. For any arm $i \in A$, we denote its online and offline means as $\mu^{\mathrm{on}}(i)=\mathbb{E}_{Y(i)\sim P_{i}^{\mathrm{on}}}[Y(i)]$, and $\mu^{\mathrm{off}}(i)=\mathbb{E}_{X(i)\sim P_{i}^{\mathrm{off}}}[X(i)]$, respectively. 

The DM aims to identify the arm with the highest mean reward in the online phase, defined as $I^{*} = \argmax_{i \in A} \mu^{\mathrm{on}}(i)$. Without loss of generality, we assume that the means of the arms are ordered such that $\mu^{\mathrm{on}}(1) > \mu^{\mathrm{on}}(2) \geq \dots \geq \mu^{\mathrm{on}}(k)$, which implies that arm 1 is the unique best arm, i.e., $I^{*} = 1$. Let $\Delta_i = \mu^{\mathrm{on}}(1) - \mu^{\mathrm{on}}(i)$ denote the suboptimality gap between arm $i$ and the best arm under the online distribution. Given $\delta > 0$, the DM seeks to design a non-anticipatory policy $\pi$ that terminates at round $\tau_{\delta} > 0$ and outputs an estimate $\hat{I}_{\tau_{\delta}}^{*}=\argmax_{i\in A}\hat{\mu}_{\tau_{\delta}}^{\mathrm{on}}(i)$, where $\hat{\mu}_{\tau_{\delta}}^{\mathrm{on}}(i)$ represents the estimate of $\mu^{\mathrm{on}}(i)$ at time $\tau_{\delta}$. The policy is required to satisfy $\mathbb{P}(\hat{I}_{\tau_{\delta}}^{*}= 1)\geq 1-\delta$, i.e., it identifies the arm with the highest online mean with probability at least $1 - \delta$. Any policy that satisfies this condition is called a $\delta$-PAC algorithm.

While existing BAI algorithms depend solely on online data and not on offline data, an ideal algorithm should be adaptive, making real-time decisions about whether to incorporate historical data in each round. If historical data is useful, integrating it can lower the sample complexity required for a correct identification probability of at least $1-\delta$. Conversely, if it is misleading, the algorithm should discard the offline data and rely solely on the online data, ensuring that its sample complexity is of the same order as the best purely online strategy. The subsequent discussion demonstrates that such an adaptive strategy is impossible without additional auxiliary information.

\section{An Impossibility Result}
Any $\delta$-PAC algorithm that utilizes historical data benefits when $P^{\mathrm{off}}$ and $P^{\mathrm{on}}$ are close, but suffers when  $P^{\mathrm{off}}$ and $P^{\mathrm{on}}$  differ significantly. However, an ideal algorithm should distinguish between these cases. In this section, we demonstrate that no ideal algorithm can exist by means of constructing lower bound on an ``alternative instance'' for any $\delta$-PAC policy. 

We consider two instances $I_{P}$ and $I_{Q}$, which share the same arm set $A = \{1,2\}$, offline sample sizes $\{T_{S}(i)\}_{i\in A}$ and the failure probability $\delta\in (0,1)$. However, $I_{P}$ and $I_{Q}$ have different reward distributions, $P$ and $Q$, and their optimal arms are also different. Specifically, for instance $I_{P}$,
\begin{align*}
    P^{\mathrm{on}}=P^{\mathrm{off}},\quad P_{1}^{\mathrm{on}}=\mathcal{N}(0,1)\quad\mathrm{and}\quad P_{2}^{\mathrm{on}}=\mathcal{N}(-\delta^{\beta},1). 
\end{align*}
Observe that the squared mean gap under $P^{\text{on}}$ satisfies $(\mu^{\text{on}}(1)-\mu^{\text{on}}(2))^{2}=\delta^{2\beta}$. If the offline sample numbers satisfy $T_{S}(1) = T_{S}(2) \geq 4(\delta^{-2\beta} - \delta^{-2\beta+\epsilon})\log(1/\delta)$, where $\epsilon\in (0,\delta)$, then, existing online policies such as Track-and-Stop \citep{agrawal2023optimal} achieves an expected stopping time at least
\begin{align}\label{P}
    &\Omega(\mathbb{E}[\tau_{\delta}']-T_{S}(1)-T_{S}(2))\notag\\
   & =\Omega(\delta^{-2\beta}\log(1/\delta)-(\delta^{-2\beta}-\delta^{-2\beta+\epsilon})\log(1/\delta))\notag\\
    &=\Omega(\delta^{-2\beta+\epsilon}\log(1/\delta)),
\end{align}
where $\mathbb{E}[\tau_{\delta}']$ is the expected stopping time without using historical data when given the error bound $\delta\in (0,1)$. Clearly, the bound in~(\ref{P}) strictly outperforms that in~(\ref{lower1}) in terms of sample complexity. Despite this apparent improvement, we show that any non-anticipatory policy that outperforms~(\ref{lower1}) on $I_{P}$ must incur a higher sample complexity than~(\ref{lower1}) on a suitably chosen $I_{Q}$. Let $\mathbb{E}[\tau_{\delta}(P)]$ and $\mathbb{E}[\tau_{\delta}(Q)]$ be the expected stopping times of the $\delta$-PAC algorithm on Instances $P$ and $Q$, respectively. The following proposition is proved in Appendix~\ref{app:proof_prop}.
\begin{proposition}\label{impossible}
    Consider instance $I_P$ as described above. Suppose the offline sample sizes satisfy $T_S(1) \in \mathbb{N}$ and there exists $\epsilon>0 $ such that $T_S(2) \leq\frac{C}{2}(\delta^{-2\beta} - \delta^{-2\beta+\epsilon})$, where $C > 0$ is an absolute constant and $\delta \in (0,1)$. Then, for any $\delta$-PAC policy $\pi$ that satisfies
    \begin{align*}
    \mathbb{E}[\tau_{\delta}(P)] \leq C\delta^{-2\beta+\epsilon}\log(1/\delta)
    \end{align*}
    on instance $I_P$, there exists an instance $I_Q$  defined as
    \begin{align*}
        &Q_1 = P_1, \quad Q_2^{\mathrm{on}} = \mathcal{N}(\delta^{\beta}, 1), \\
        &Q_2^{\mathrm{off}} = \mathcal{N}\bigg(-\sqrt{\frac{8\delta^{-2\beta - \epsilon}}{C(1 - \delta^{\epsilon})}} - \delta^{\beta}, 1\bigg),
    \end{align*}
    such that as $\delta\to 0^+$, the policy $\pi$ suffers
    \begin{align*}
    \mathbb{E}[\tau_{\delta}(Q)]\geq \Omega\bigg(\frac{1}{\delta^{2\beta+\epsilon}}\log \Big(\frac{1}{\delta}\Big)\bigg)=\omega(\mathbb{E}[\tau_{\delta}']). 
    \end{align*}    
\end{proposition}

We first analyze instances $I_{P}$ and $I_{Q}$. In instance $I_{P}$, arm~1 is the better arm, while in instance $I_{Q}$, arm 2 is the better arm. Instances $I_{P}$ and $I_{Q}$ share the same arm set, same offline sample sizes which are sufficiently large to reduce the sample complexity on instance $I_P$ with parameter $\delta$. Although the online reward distributions are different, with $P_2^{\mathrm{on}} = \mathcal{N}(-\delta^\beta, 1)$ in $I_P$ and $Q_2^{\mathrm{on}} = \mathcal{N}(\delta^\beta, 1)$ in $I_Q$, a sign flip and arm relabeling make the two instances equivalent to any online-only policy. The offline datasets 
are generated from different offline reward distributions. Consequently, offline data plays a critical role in these instances: it can be helpful in $I_{P}$, but misleading in $I_{Q}$. Without additional information, we cannot determine whether the offline data will be helpful or misleading.  


Proposition \ref{impossible} shows that an ideal algorithm does not exist. Even when we consider an extreme case when $T_{S}(1)=T_{S}(2)=\infty$, which implies that the DM knows the exact mean value of the offline reward distributions $P^{\mathrm{off}}$, it is not possible to find an algorithm simultaneously (1) if the offline data is helpful (instance $I_{P}$), it could reduce sample complexity compared to pure online strategies like LUCB; (2) if the offline data is misleading (instance $I_{Q}$), it should be ``smart'' enough to discard the historical data and match the sample complexity of the standard LUCB algorithm. 

In the following section, we introduce an LUCB-based algorithm, called LUCB-H, which incorporates an additional input: the valid biased bound $V=\{V(i)\}_{i\in A}\in(\mathbb{R}\cup \{\infty\})^{k}$ on an instance $I$ such that
\begin{equation*}
V(i)\geq |\mu^{\text{off}}(i)-\mu^{\text{on}}(i)|,\quad\text{for each $i\in A$}.
\end{equation*}
The quantity $V(i)$ serves as an upper bound on the mean shift between the offline and online reward distributions $P^{\mathrm{off}}(i)$ and $P^{\mathrm{on}}(i)$ for arm $i$. Before the online phase, the DM can construct the prior knowledge of $V(i)$ by several machine learning estimators, such as the LASSO \citep{blanchet2019robust}, by cross-validation \citep{chen2022data}, or  obtain some insights on how to construct them empirically \citep{si2023distributionally}.

If $V(i)=\infty$, then the DM has no prior knowledge about the difference between the online and offline distributions before the online phase. In this case, Proposition \ref{impossible} has already demonstrated that no ideal algorithm exists. Therefore, we assume that $V(i)<\infty$. Similar upper bound can be found in the contexts of supervised learning \citep{crammer2008learning}, offline policy learning on contextual bandits \citep{si2023distributionally}, stochastic optimization \citep{besbes2022beyond} and multi-task bandit learning \citep{wang2021multitask}.

\begin{algorithm}[H]
\caption{LUCB-H Algorithm}
\textbf{Input:} Valid bias bound $\{V(i)\}_{i=1}^{k}$ on the instance, confidence parameter $\delta$, offline samples $S$

\begin{algorithmic}[1]
\STATE For each $i \in A$, compute $\hat{X}(i)$ and set $\hat{Y}_0(i) = 0$
\STATE At $t = 1, \ldots, k$, pull each arm once, then set $N_{k+1}(i) = 1$ for all $i\in A$
    \FOR{$t=k+1, k+2,\ldots$}
        \STATE \textcolor{black}{// Construct   lower and upper bounds.} 
        \STATE Compute $\mathrm{LCB}_{t}(i)$ and $\mathrm{UCB}_{t}(i)$ by (\ref{lon})--(\ref{uon})
        \STATE Compute $\mathrm{LCB}_{t}^{S}(i)$ and $\mathrm{UCB}_{t}^{S}(i)$ by~(\ref{loff})--(\ref{uoff})
        \STATE Compute $\mathrm{LCB}^{\mathrm{mix}}(i)=\max\{\mathrm{LCB}_{t}(i), \mathrm{LCB}_{t}^{S}(i)\}$ and $\mathrm{UCB}^{\mathrm{mix}}(i)=\min\{\mathrm{UCB}_{t}(i), \mathrm{UCB}_{t}^{S}(i)\}$
        
        \STATE Set $h_{t}=\argmax_{i\in A}\mathrm{UCB}^{\mathrm{mix}}(i)$
        \STATE Set $l_{t}=\argmax_{i\neq h_{t}}\mathrm{UCB}^{\mathrm{mix}}(i)$
    \ENDFOR
    \IF{ $\mathrm{LCB}^{\mathrm{mix}}(h_t) \geq \mathrm{UCB}^{\mathrm{mix}}(l_t)$ }
    \STATE Break the loop.
    \ENDIF
    \STATE Pull arms $h_{t}$ and $l_{t}$, and observe independent rewards: $Y_t(h_{t}) \sim P^{\mathrm{on}}_{h_t}$ and $Y_t(l_{t}) \sim P^{\mathrm{on}}_{l_t}$. 
    \STATE Update sample mean $\hat{Y}_{t+1}(i)$ and sample counters $N_{t+1}(i)$ by
    \begin{align*}
        \hat{Y}_{t+1}(i) = 
            \begin{cases}
                \frac{N_t(i) \cdot \hat{Y}_t(i) + Y_t(i)}{N_t(i) + 1}, & \text{if } i = h_{t} \mbox{ or } l_{t} \\
                \hat{Y}_t(i), & \text{otherwise},
            \end{cases}
    \end{align*}
    and $N_{t+1}(i) = N_t(i)+\mathbbm{1}\{i\in \{h_{t}, l_{t}\}\}$.
\end{algorithmic}
\textbf{Output:} \( I^{*}=\argmax_{i\in A}\mathrm{UCB}^{\mathrm{mix}}(i) \)
\end{algorithm}

\section{Design and Analysis of the LUCB-H Algorithm}
The LUCB algorithm was first proposed by \citet{kalyanakrishnan2012pac}. Unlike the UCB algorithm, which is more suitable for cumulative reward maximization, LUCB is specifically designed for the BAI problem (more precisely, the best $m$-arm identification problem where $m\in A$). In each round, LUCB selects two arms: the one with the highest sample mean and the one with the largest upper confidence bound among the remaining arms. LUCB provides a theoretical upper bound on the error probability under the PAC framework and achieves near-optimal sample complexity in many cases. As an attractive approach for finite-action stochastic bandits, it has been 
extended to other variants, such as the top feasible arm identification problem \citep{katz2019top}. With these merits in mind, it seems quite natural to generalize the idea of LUCB to address the BAI problem in a historical data setting.

To extend the LUCB algorithm to a setting with historical data, the main challenge is deciding how much to rely on offline data while ensuring accurate online decision-making. To address this problem, LUCB-H computes both the upper and lower confidence bounds using two types of estimators.
\begin{itemize}[leftmargin=*]
    \item The first uses \textbf{only online data} and follows the standard LUCB method:
    \begin{align}
    \mathrm{LCB}_{t}(i) &= \hat{Y}_t(i) - \sqrt{\frac{2 \log(kt / \delta)}{N_t(i)}}\label{lon} \quad \mbox{and}\\
    \mathrm{UCB}_{t}(i) &= \hat{Y}_t(i) + \sqrt{\frac{2 \log(kt / \delta)}{N_t(i)}},\label{uon}
\end{align}
where \(\hat{Y}_t(i)\) is the empirical mean of arm \(i\).
\item The second \textbf{incorporates historical data}. Let $\hat{X}(i)$ be the historical mean, i.e., $\hat{X}(i)=\frac{\sum_{s=1}^{T_S(i)} X_s(i)}{T_S(i)}$. Then at round $t$, the estimated mean is
\begin{equation*}\label{his_mean}
    \hat{Y}_{t}^{S}(i) = \frac{N_t(i) \cdot \hat{Y}_t(i) + T_S(i) \cdot \hat{X}(i)}{N_t(i) + T_S(i)}, 
\end{equation*}
with adjusted confidence bounds 
\begin{align}
    \mathrm{LCB}_{t}^{S}(i) &= \hat{Y}_t^S(i) - \mathrm{rad}_t^S(i)\label{loff}\quad \mbox{and}\\
    \mathrm{UCB}_{t}^{S}(i) &= \hat{Y}_t^S(i) + \mathrm{rad}_t^S(i),\label{uoff}
\end{align}
respectively, where
\begin{align*}
    \mathrm{rad}_t^S(i) = \sqrt{\frac{2 \log(kt / \delta)}{N_t(i) + T_S(i)}} + \frac{T_S(i)}{N_t(i) + T_S(i)} \cdot V(i).
\end{align*}
\end{itemize}
The above confidence interval consists of two parts: the first term is a standard confidence bound after incorporating historical data, and the second term compensates for potential bias, which is quantified by the parameter $V(i)$. 

The core idea of the LUCB algorithm is to iteratively narrow down the set of candidate arms using these confidence bounds, which accelerates convergence. Inspired by this insight, the algorithm then conservatively selects the tighter bounds, i.e., 
\begin{align*}
      \mathrm{LCB}_{t}^{\mathrm{mix}}(i)&=\max\{\mathrm{LCB}_{t}(i), \mathrm{LCB}_{t}^{S}(i)\},\quad \mbox{and}\\
    \mathrm{UCB}_{t}^{\mathrm{mix}}(i)&=\min\{\mathrm{UCB}_{t}(i), \mathrm{UCB}_{t}^{S}(i)\}.
\end{align*}
Since the arm with the largest sample mean may differ with or without historical data, LUCB-H selects $h_t$ as the arm with the largest upper bound, then proceeds as in standard LUCB.


Next, we analyze $\mathrm{UCB}_t^S(i)$ and $\mathrm{UCB}_t(i)$. When the bias bound $V(i)$ is small and the offline sample size $T_S(i)$ is large, typically $\mathrm{UCB}_t^S(i)<\mathrm{UCB}_t(i)$, indicating historical data can reduce confidence intervals and speed up convergence. A small $V(i)$ suggests similarity between $P^{\mathrm{on}}$ and $P^{\mathrm{off}}$, making historical data beneficial; otherwise, historical data should be discarded.

In addition, let us compare the expectations of $\mathrm{UCB}_{t}^S(i)$ and $\mathrm{UCB}_{t}(i)$, to intuitively analyze the impact of the parameter $V(i)$ and the amount of historical data for arm $i \in A$ on the algorithm's performance. Here, we focus on round $t$ and, for the sake of exposition,  we treat $N_t(i)$ as a deterministic quantity, which we denote as $n_t(i)$. Then, 
\begin{align*}
    &\mathbb{E}[\mathrm{UCB}_{t}^S(i)|N_{t}(i)=n_{t}(i)]\\
    &= \frac{n_t(i) \cdot\mu^{\mathrm{on}}(i) + T_S(i) \cdot \mu^{\mathrm{off}}(i)}{n_t(i) + T_S(i)}+\frac{T_S(i)}{n_t(i) + T_S(i)} \cdot V(i)\\
    &\quad+\sqrt{\frac{2 \log(kt / \delta)}{n_t(i) + T_S(i)}}\\
    &=\mu^{\mathrm{on}}(i)+\frac{T_S(i)}{n_t(i) + T_S(i)} (V(i)+\mu^{\mathrm{off}}(i)-\mu^{\mathrm{on}}(i))\\
    &\quad+\sqrt{\frac{2 \log(kt / \delta)}{n_t(i) + T_S(i)}}
\end{align*}
and 
\begin{align*}
    \mathbb{E}[\mathrm{UCB}_{t}(i)|N_{t}(i)=n_{t}(i)]=\mu^{\mathrm{on}}(i)+\sqrt{\frac{2 \log(kt / \delta)}{n_t(i)}}.
\end{align*}


As analyzed above, the biased bound $V(i)$ and the actual mean shift between offline and online data $\mu^{\mathrm{off}}(i)-\mu^{\mathrm{on}}(i)$ determine whether the historical data is helpful for identifying the best arm, and the number of offline samples $T_S(i)$ plays a critical role in the reducing sample complexity. Hence, we define a discrepancy measure
\begin{align*}
    \eta(i)=V(i)+\mu^{\mathrm{off}}(i)-\mu^{\mathrm{on}}(i)
\end{align*}
on arm $i$ and we can know that $\eta(i)\in [0,2V(i)]$. When the length of the confidence interval on arm $i$ is smaller than $\Delta_i/4$, sampling for arm $i$ can stop \citep{jamieson2014best}. Therefore, we can infer that when $\eta_i < \Delta_i/4$, the historical data for arm $i$ is helpful for the identification process. The following theorem further supports this inference.

\begin{theorem}\label{thm: 4.1}
    The LUCB-H algorithm, which inputs a valid bias bound $V$ on instance $I$, the expected stopping time $\mathbb{E}[\tau_{\delta}]$ satisfies
    \begin{align}\label{upper bound}
        \mathbb{E}[\tau_{\delta}]=O\bigg(\sum_{\Delta_{i}>0}\bigg(\frac{1}{\Delta_{i}^{2}}\log(\frac{1}{\delta})-\mathrm{Sav}_{\mathrm{u}}(i)\bigg)\bigg),
    \end{align}
    where the ``Saving'' term in the upper bound is $$\mathrm{Sav}_{\mathrm{u}}(i)=T_{S}(i)\cdot\max\bigg\{1-\frac{4\eta(i)}{\Delta_{i}},0 \bigg\}.$$ 
\end{theorem}
Theorem \ref{thm: 4.1} is provided in   Appendix~\ref{app:proof_ach}.
The upper bound in (\ref{upper bound}) is not greater than the expected stopping time in the standard case using LUCB (without utilizing historical data).
Note that $\mathrm{Sav}_{\mathrm{u}}(i) \geq 0$. 
When $\eta(i) \leq \Delta_i/4$, $P^{\mathrm{on}}(i)$ and $P^{\mathrm{off}}(i)$ are sufficiently close, indicating that the historical data is helpful, and the sample complexity of LUCB-H can be \emph{strictly less} than that of LUCB. Otherwise, the historical data is misleading, and the DM should discard it. In this case, LUCB-H adaptively \emph{retains} the sample complexity of the standard LUCB algorithm. If $\eta(i) \leq \Delta_i/4$, then $\mathrm{Sav}_{\mathrm{u}}(i)$ is monotonically increasing with respect to $T_S(i)$. This means that the more helpful historical data we have, the better it is for identifying the best arm, which is consistent with common intuition.

\section{Lower Bound of Any $\delta$-PAC Algorithm}

This section characterizes the complexity of the BAI problem when using possibly biased offline data. It assumes that the set of probability measures \(\mathcal{P}\) satisfies a common assumption used in the BAI and MAB literature \citep{lai1985asymptotically, kaufmann2016complexity}, specifically related to the continuity of the KL divergence. This assumption enables us to develop change-of-measure arguments for deriving lower bounds on the problem's complexity.

\begin{assumption}\label{continuity}
For all $v, v' \in \mathcal{P}^2$ such that  $v \neq v'$, for all $a > 0$, there exists 
$ v'' \in \mathcal{P} $ such that $ \mathrm{KL}(v, v') < \mathrm{KL}(v, v'') < \mathrm{KL}(v, v') + a$ and  $\mathbb{E}_{X \sim v''}[X] > \mathbb{E}_{X \sim v'}[X]
$. Furthermore, there exists $v''' \in \mathcal{P}$ such that $ \mathrm{KL}(v, v') < \mathrm{KL}(v, v''') < \mathrm{KL}(v, v') + a$ and $\mathbb{E}_{X \sim v'''}[X] < \mathbb{E}_{X \sim v'}[X]$.
\end{assumption}

\begin{theorem}\label{thm 5.1}
Suppose that $\mathcal{P}$ satisfies Assumption \ref{continuity}; any algorithm that is $\delta$-PAC on $\mathcal{M}$ satisfies, for $\delta \leq 0.15$, the expected stopping time
\begin{align*}
    \mathbb{E}[\tau_{\delta}]\!=\!\Omega\bigg( \sum_{\Delta_{i}>0}\! \! \bigg(\frac{1}{\mathrm{KL}(\mu^{\mathrm{on}}(i), \mu^{\mathrm{on}}(1))}  \log\big(\frac{1}{\delta}\big)\!-\!\mathrm{Sav}_{\mathrm{l}}(i)\bigg)\!\bigg),
\end{align*}
where $\mathrm{Sav}_{\mathrm{l}}(i)$, the   ``Saving'' term in the lower bound, is
\begin{align*}
    \mathrm{Sav}_{\mathrm{l}}(i)=T_{S}(i)\max\bigg\{\frac{\mu^{\mathrm{off}}(1) -\mu^{\mathrm{off}}(i) }{\Delta_{i}},0\bigg\}^{2}.
\end{align*}
\end{theorem}
Comparing the lower bound to  standard LUCB, LUCB-H improves the bound on the   stopping time bound by including  the saving term $\mathrm{Sav}_{\mathrm{l}}(i)$ for arms $i\in A$. The proof of Theorem \ref{thm 5.1} can be found in  Appendix~\ref{app:proof_conv}.

Next, we will analyze the upper bound of the stopping time of the LUCB-H algorithm and the lower bound of BAI with biased historical data. Clearly, the upper bound of LUCB-H is almost equal to  the lower bound of sample complexity for the BAI problem with possibly biased offline data. Specifically, when the offline data are misleading and identical, the two bounds are identical. In the following, we quantify the gap between two bounds in two cases when the saving terms $\mathrm{Sav}_{\mathrm{u}}(i)>0$ and $\mathrm{Sav}_{\mathrm{l}}(i)>0$. We define the {\em gap} between them as $\mathrm{gap}(i) := \mathrm{Sav}_{\mathrm{l}}(i)-\mathrm{Sav}_{\mathrm{u}}(i)$.

\begin{remark}
    We analyze the above-defined gap  $\mathrm{gap}(i)$ in two special cases. 
    \begin{enumerate}[leftmargin=*]
        \item $V(i)$'s are equal:  In this case, we can show that  gap is non-negative because 
        \begin{align*}
            \mathrm{gap}(i) \!=\! \bigg( \frac{ (\eta({i})-\eta({1}))^{2}}{\Delta_{i}^{2}}+2\frac{\eta({i})+\eta({1})}{  \Delta_{i} }\bigg) T_{S}(i)\ge 0.
        \end{align*}

        \item $V(1)=\eta( 1)$, i.e., $P^{\mathrm{on}}(1)=P^{\mathrm{off}}(1)$. In this case, we can also show that the gap is non-negative because 
             \begin{align*}
            \mathrm{gap}(i) \!=\! \bigg(\frac{(\eta({i})-V(i))^{2}}{\Delta_{i}^{2}}+2\frac{\eta({i})+V(i)}{\Delta_{i}}\bigg) T_{S}(i)\ge 0.
        \end{align*}
        While we believe that $\mathrm{gap}(i)\ge0$ holds generally, we leave the proof of this to future work. 
    \end{enumerate}
\end{remark}


\begin{figure*}[t]
    \centering
    \begin{minipage}{0.32\textwidth} 
        \centering
        \includegraphics[width=\textwidth]{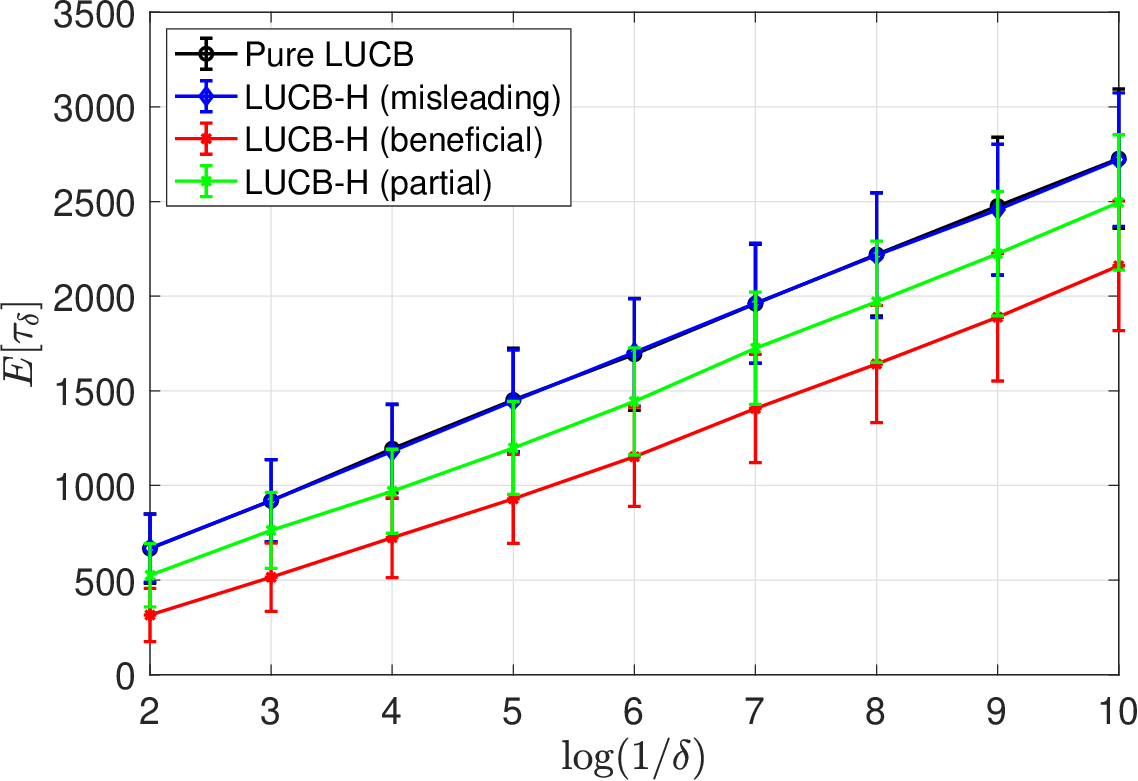}
        \caption{Evolution of $\mathbb{E}[\tau_{\delta}]$ with $\log(1/\delta)$ in group $1$}
        \label{fig:fig1}
        \includegraphics[width=\textwidth]{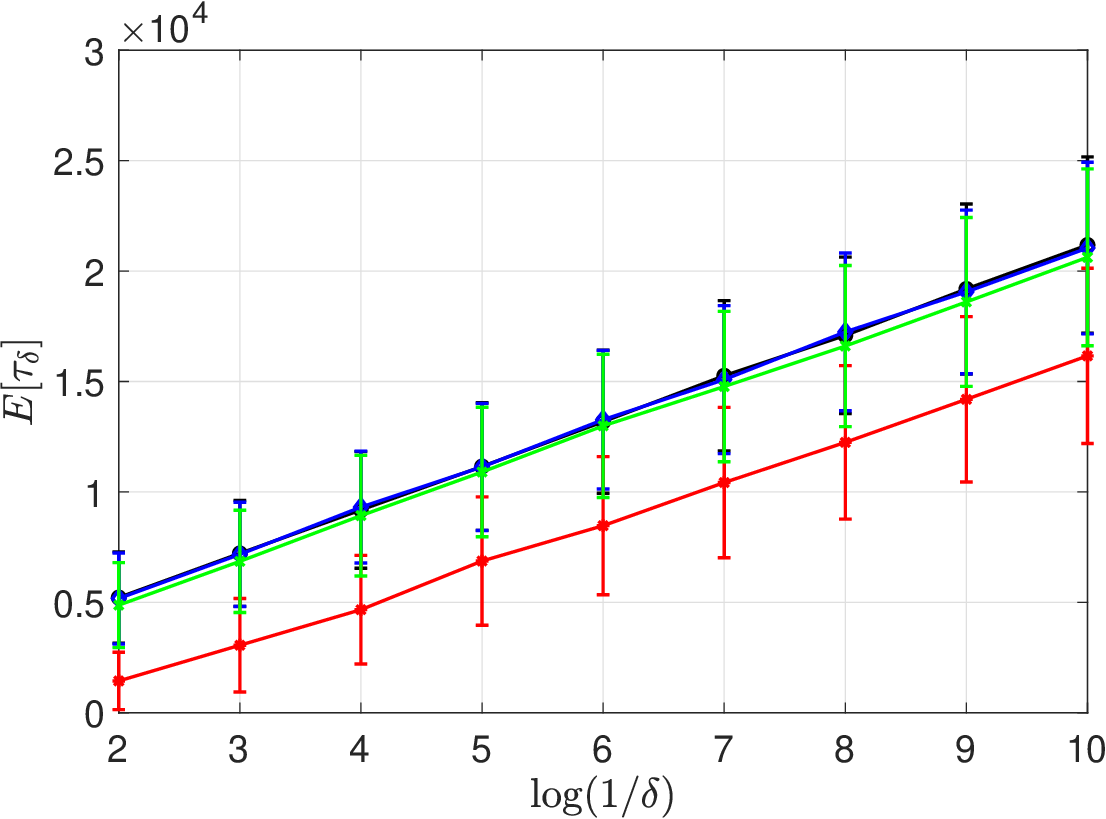}
        \caption{Evolution of $\mathbb{E}[\tau_{\delta}]$ with $\log(1/\delta)$ in group $2$}
        \label{fig:fig2}
    \end{minipage}
    \hfill  
    \begin{minipage}{0.34\textwidth} 
        \centering
        \includegraphics[width=\textwidth]{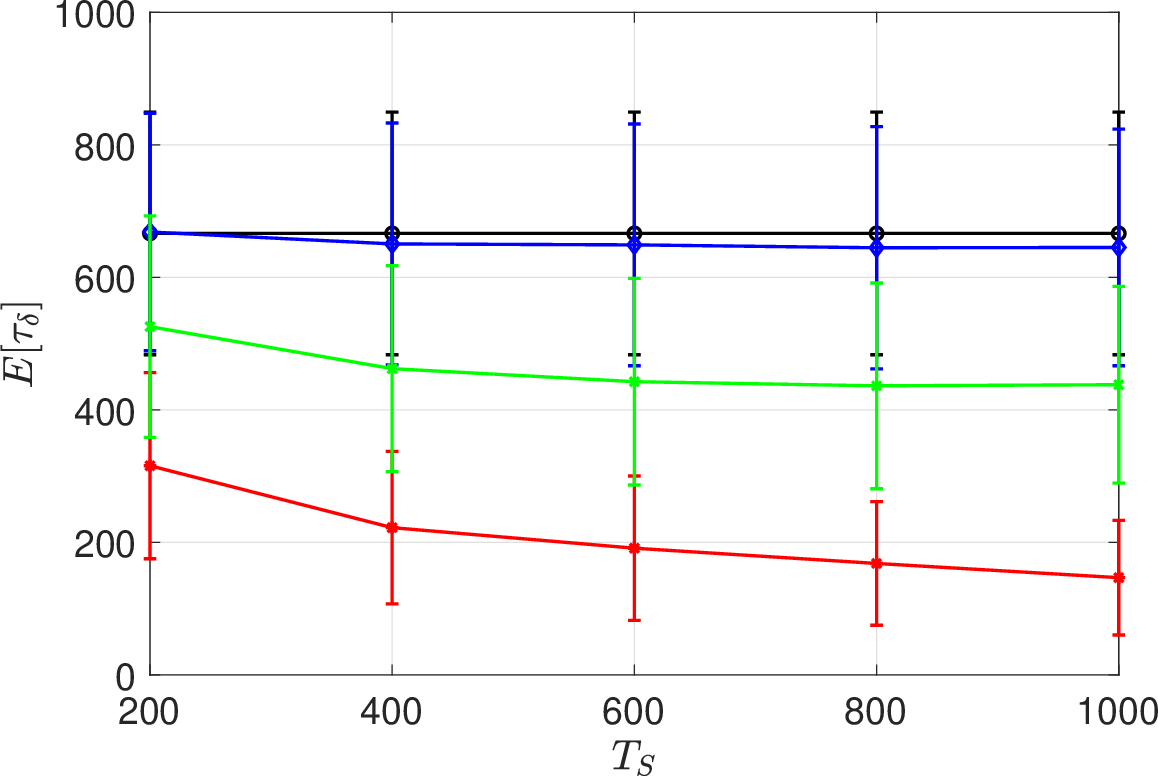}
        \caption{Evolution of $\mathbb{E}[\tau_{\delta}]$ with $T_{S}$ in group $1$}
        \label{fig:fig3}
        \includegraphics[width=\textwidth]{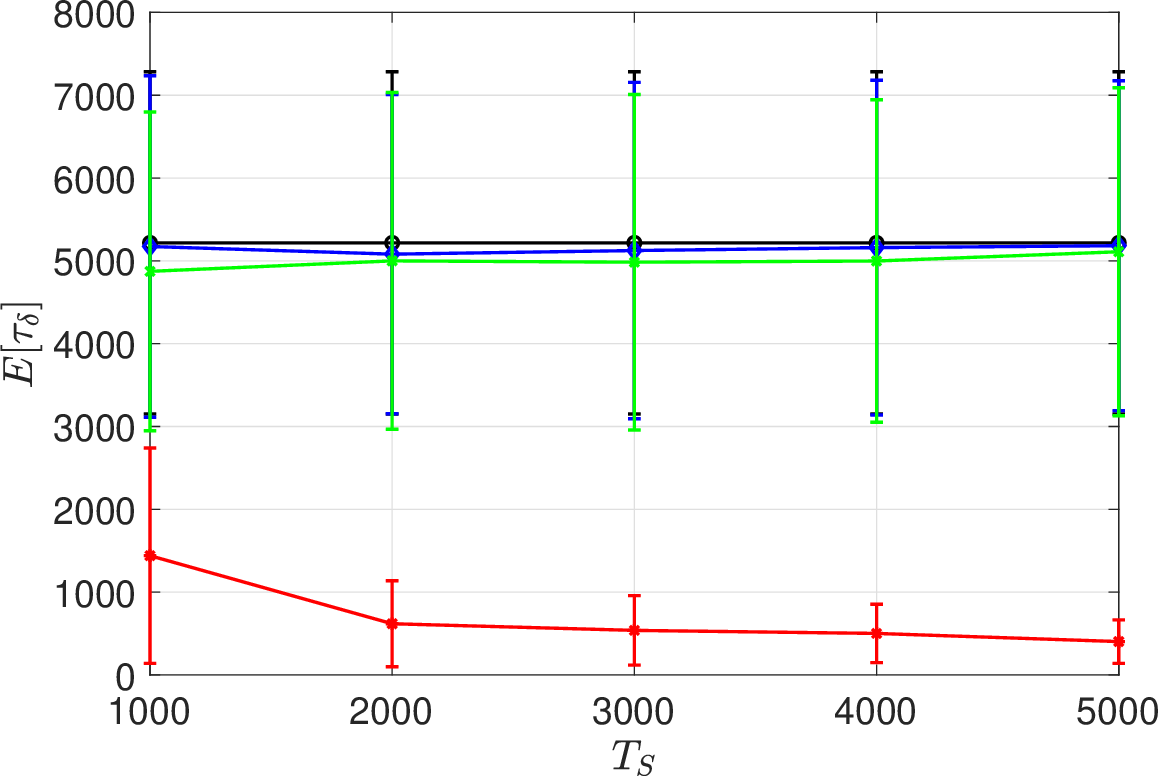}
        \caption{Evolution of $\mathbb{E}[\tau_{\delta}]$ with $T_{S}$ in group $2$}
        \label{fig:fig4}
    \end{minipage}
    \hfill  
    \begin{minipage}{0.33\textwidth} 
        \centering
        \includegraphics[width=\textwidth]{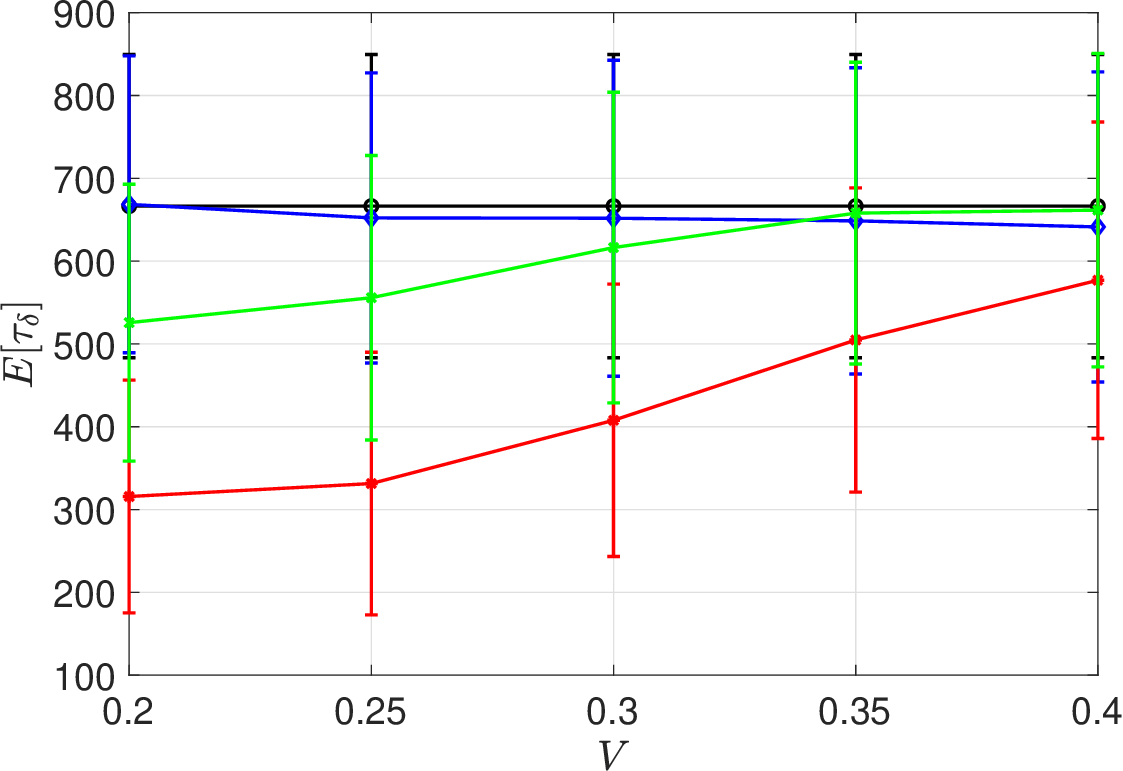}
        \caption{Evolution of $\mathbb{E}[\tau_{\delta}]$ with $V$ in group $1$}
        \label{fig:fig5}
        \includegraphics[width=\textwidth]{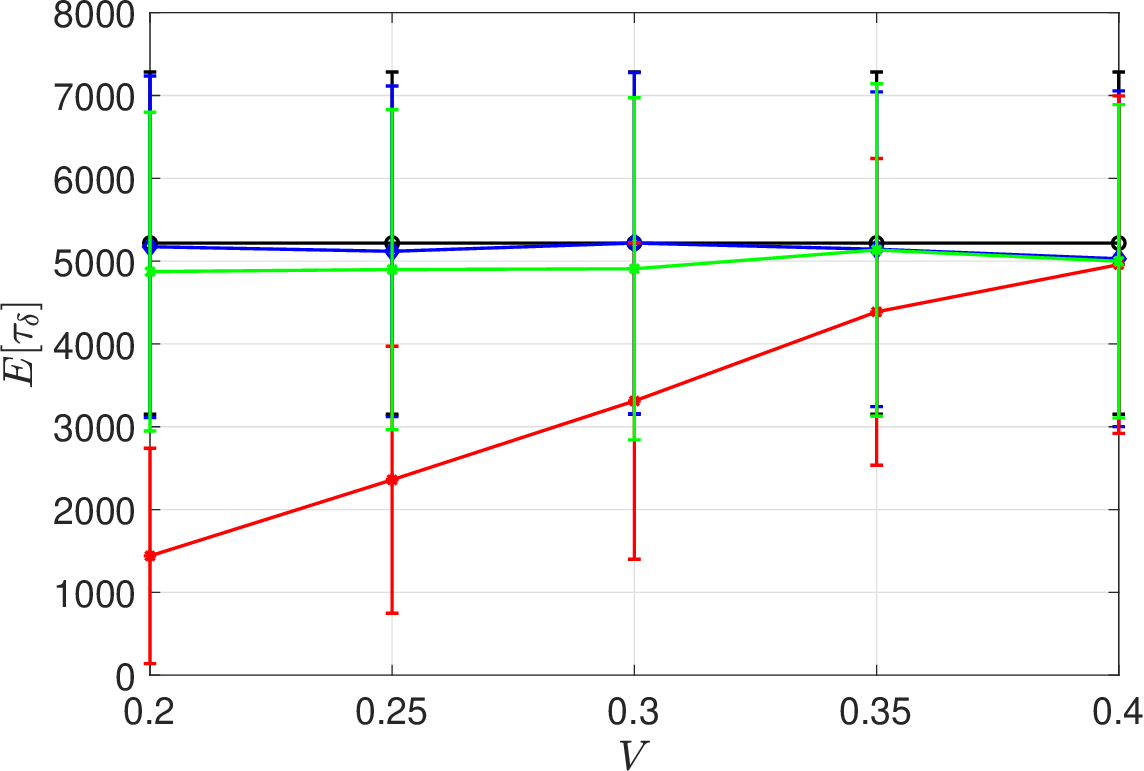}
        \caption{Evolution of $\mathbb{E}[\tau_{\delta}]$ with $V$ in group $2$}
        \label{fig:fig6}
    \end{minipage}
\end{figure*}

\begin{figure*}[t]
    \centering
    \begin{minipage}{0.42\textwidth}
        \centering
        \includegraphics[width=\textwidth]{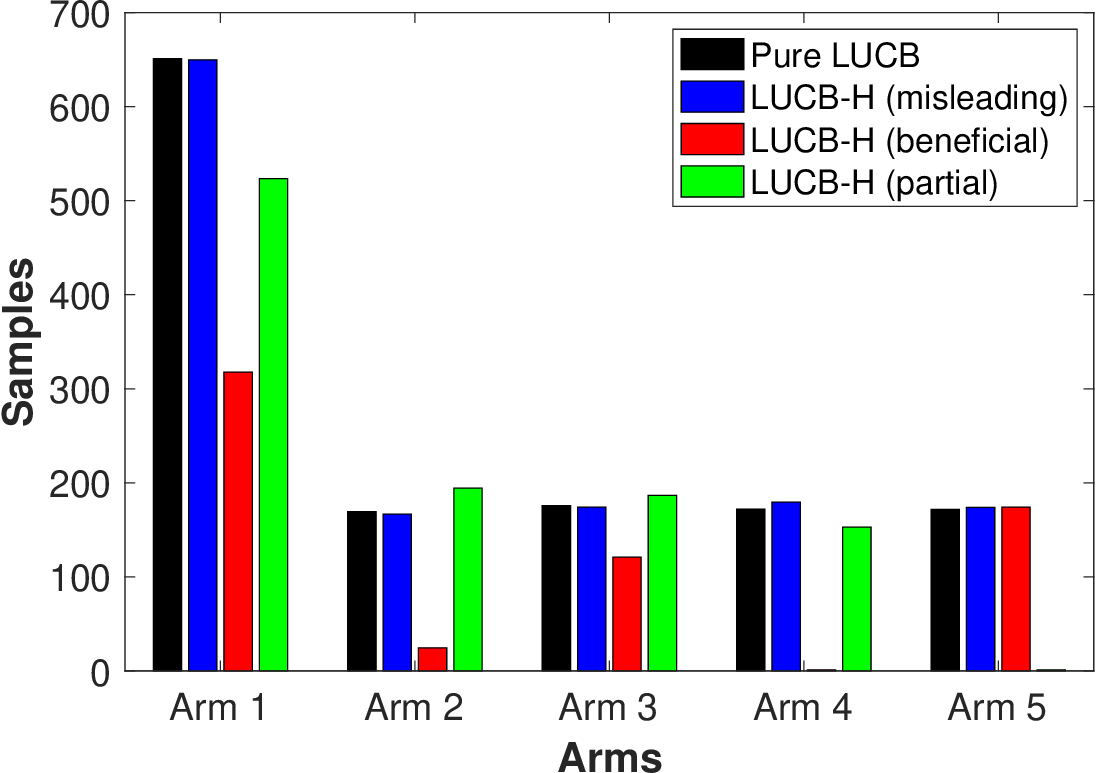}
        \caption{Samples for different arms when $\delta = 0.01$ in group $1$}
        \label{fig:fig7}
    \end{minipage}
    \hspace{0.04\textwidth}
    \begin{minipage}{0.42\textwidth}
        \centering
        \includegraphics[width=\textwidth]{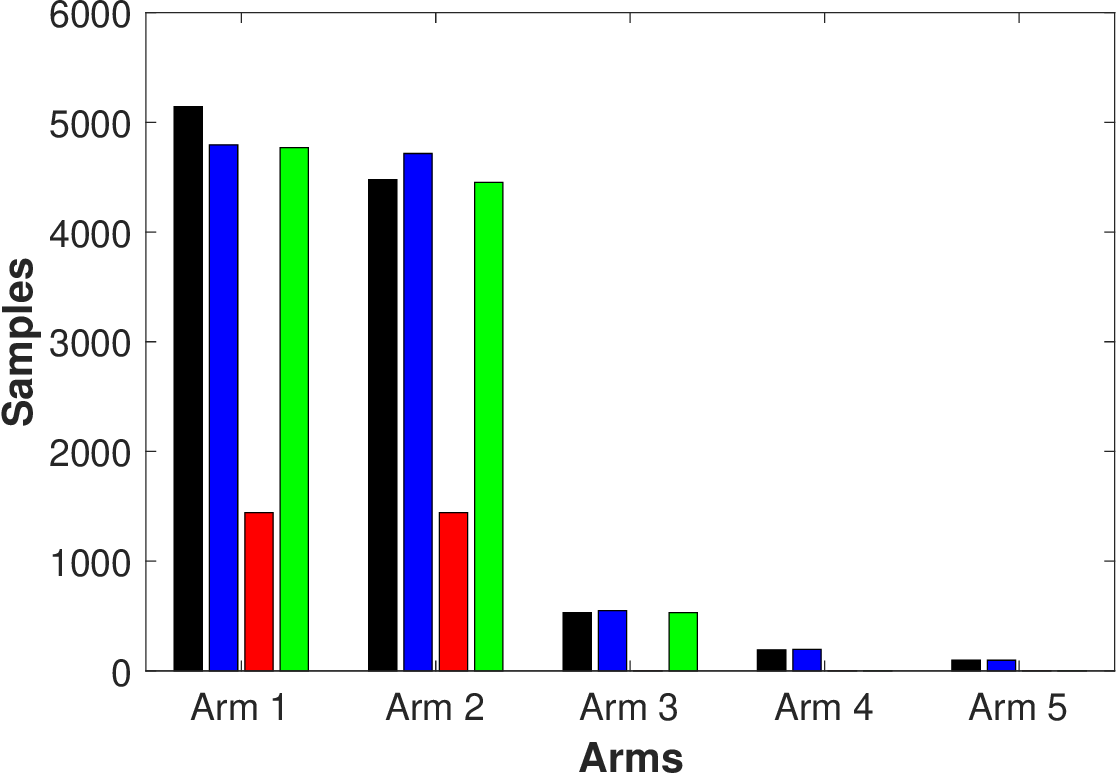}
        \caption{Samples for different arms when $\delta = 0.01$ in group $2$}
        \label{fig:fig8}
    \end{minipage}
\end{figure*}

\section{Numerical Results}
In this section, we conducted numerical experiments to evaluate the performance of the LUCB-H algorithm, which leverages biased offline data, and compared its performance with the Pure LUCB algorithm (the LUCB algorithm, which does not utilize any historical data). All experiments were based on a 5-armed bandit problem with rewards corrupted by Gaussian noise with unit variance, and the results reported are averaged over $1000$ trials. The overall experimental design is divided into two groups based on the online reward distributions (staircase and linear), with both groups incorporating the same offline data construction and experimental variable settings. 

We will introduce the design of online and offline reward distributions first. In Group $1$, the online reward distributions exhibit a staircase pattern with means of $(0.8, 0.4, 0.4, 0.4, 0.4)$. In Group $2$, the online reward distributions follow a linear pattern with means of $(0.8, 0.7, 0.6, 0.5, 0.4)$. Note that in both groups, arm $1$ is the best arm, and the Pure LUCB algorithm makes its decision solely based on online data.

For each group, three different offline reward distributions are constructed (with the offline data generated from $P^{\mathrm{off}}$ and with different number of historical samples per arm):
\begin{enumerate}[leftmargin=*,label=(\alph*)]
	\item \textbf{Case $1$ (Misleading Bias):} In this case, all arms are adversely affected. Thus, the offline mean for the best arm  decreases while the means for the suboptimal arms increase. The offline means in Group $1$ are $(0.4, 0.6, 0.6, 0.6, 0.6)$,
		resulting in the offline phase identifying a best arm that is not arm $1$. The offline means in Group $2$ are $(0.4, 0.8, 0.7, 0.6, 0.5)$, which likewise leads to an offline best arm being different from arm~$1$.
	
	\item \textbf{Case $2$ (Beneficial Bias):} In this case, the offline data assists in the identification process, with the offline mean for the best arm being increased while those for the non-best arms are decreased. The offline means in Group $1$ are $(0.9, 0.2, 0.2, 0.2, 0.2)$ so that the offline phase still correctly identifies arm $1$ as the best. The offline means in Group $2$ are $(0.9,  0.5,  0.4,  0.3, 0.2)$, where the offline best arm remains arm $1$.
	
	\item \textbf{Case 3 (Partial Adjustment):} In this case, selective adjustments are made across the arms. The offline means in Group $1$ are $(0.4, 0.6, 0.6, 0.2, 0.2)$, under which the offline phase will not select arm $1$ as the best. The offline means in Group $2$ are $(0.4, 0.8, 0.7, 0.3,0.2)$, leading to an offline best arm different from arm $1$.
\end{enumerate}

We also explored the impact of $T_{S}(i)$ and $V(i)$ on the performance of the LUCB-H algorithm. The two factors represent the amount of offline data and the deviation between offline and online data, respectively, which can significantly affect the convergence speed and the accuracy of identifying the best arm. We assess the expected stopping times of Pure LUCB and LUCB-H  under three   instances:

\begin{enumerate}[leftmargin=*]
	\item \textbf{Variation of $\delta$:} The offline sample size is fixed at $T_{S}(i) = 200$ in Group $1$ and $T_{S}(i) = 1000$ in Group $2$ for all $i \in A$, and the bias parameter is configured as $V = (0.4, 0.2, 0.2, 0.2, 0.2)$ in Group $1$ and Group $2$. We set the confidence levels at $\delta=0.1, 0.1^2, \dots, 0.1^{10}$. 
	
	\item \textbf{Variation of Offline Sample Size $T_{S}(i)$ and Parameter $V(i)$:} Under a fixed $\delta=0.01$:
	\begin{itemize}[leftmargin=*]
		\item The offline sample size for each arm is varied among $200$, $400$, $600$, $800$ and $1000$ in Group $1$ and among $1000$, $2000$, $3000$, $4000$ and $5000$ in Group $2$. The bias parameter in these experiments are also $V = (0.4, 0.2, 0.2, 0.2, 0.2)$.
		\item When the offline sample size is fixed at $200$ for each arm, the effect of varying parameter $V$ (fixed $V(1)=0.4$ and $V(i)$ ranging from $0.2$ to $0.4$ for all suboptimal arms in two groups) on the expected stopping time is further examined.
	\end{itemize}
\end{enumerate}

Figures \ref{fig:fig1} to \ref{fig:fig6} analyze how the average stopping time $\mathbb{E}[\tau_\delta]$ of the LUCB algorithm, and the LUCB-H algorithm under misleading, beneficial, and partial instances varies from three perspectives: $\log(1/\delta)$, offline sample size $T_S$, and parameter $V$. Across all experiments, LUCB-H behaves similarly to Pure LUCB in the misleading instance, both in terms of mean stopping time and error bars, regardless of changes in confidence level $\delta$, offline sample size $T_S(i)$, or biased bound $V(i)$. This suggests that LUCB-H wisely discards misleading offline data. In the partial instances, LUCB-H's performance consistently lies between that of the beneficial and misleading instances. The following analysis focuses on the beneficial instance.

Figures \ref{fig:fig1} and \ref{fig:fig2} show how $\mathbb{E}[\tau_\delta]$ evolves with $\log(1/\delta)$. All curves exhibit linear or near-linear growth.  The slope of each curve approximates the sample complexity of the corresponding algorithm. LUCB-H achieves sample complexity that matches that of LUCB in general. LUCB-H requires significantly fewer samples in the beneficial instance compared to Pure LUCB and LUCB-H in other instances. As $\delta \to 0$, the error bars in the beneficial instance gradually expand and converge with those of Pure LUCB, indicating a diminishing influence of historical data.

Figures \ref{fig:fig3} and \ref{fig:fig4} illustrate  how the stopping time varies with offline sample size $T_S$. In the beneficial instance, the stopping time decreases sharply as $T_S$ increases, demonstrating the advantage of helpful offline data. The error bars also shrink with increasing $T_S$, indicating that even biased beneficial data can effectively narrow the confidence bounds and accelerate convergence.


Figures \ref{fig:fig5} and \ref{fig:fig6} examine how the parameter \(V\) influences the expected stopping time. In the beneficial instance, the stopping time increases notably as \(V\) grows. This indicates that, even when historical data is advantageous, a large bound on the shift between online and offline reward distributions can adversely impact LUCB-H's performance, revealing its sensitivity to substantial shifts. Although the error bars in the beneficial case widen with increasing \(V\), potentially affecting reliability, LUCB-H still performs comparably to Pure LUCB in this scenario.


Figures \ref{fig:fig7} and \ref{fig:fig8} summarize the sample allocation per arm for both Pure LUCB and LUCB-H in Groups 1 and 2 when \(\delta = 0.01\). 
In the misleading case, LUCB-H assigns samples similarly to Pure LUCB, demonstrating its ability to avoid relying on misleading historical data and behave accordingly. 
In the beneficial instance, LUCB-H allocates significantly fewer samples per arm compared to Pure LUCB, illustrating its effectiveness in leveraging helpful historical information. 
In the partial scenario, LUCB-H dynamically adapts its sampling; it assigns substantially fewer samples to arms with beneficial offline rewards, while allocating a similar number as Pure LUCB to arms with misleading offline data. This showcases LUCB-H's flexibility in selectively utilizing historical data on an arm-by-arm basis. 
Additionally, some arms receive only one sample, as LUCB concentrates most samples on the top two arms. When both offline and initial online data strongly suggest an arm is suboptimal, LUCB-H may cease sampling further, reflecting its adaptivity. In Appendix~\ref{app:robustnes}, we examine the robustness of LUCB-H to misspecifications in the bias bound $V(i)$.

\vspace{-.18em}
\section{Conclusions}\vspace{-.18em}
In this paper, we studied the best arm identification problem with potentially biased offline data. We first established an impossibility result, proving that no $\delta$-PAC algorithm can out-perform the state-of-the-art on all instances, without prior knowledge of the bias bound between offline and online distributions. This result highlights the necessity of incorporating auxiliary knowledge, such as bias bounds, to effectively utilize offline data while ensuring robust performance. To address this challenge, we proposed the LUCB-H algorithm. The algorithm adaptively balances offline and online data by calculating upper and lower confidence bounds separately with and without offline data for each arm. It selects the smaller upper confidence bound and larger lower confidence bound to obtain more accurate mean estimates. This design allows LUCB-H to decide how much to rely on offline data at each iteration. Our theoretical analysis shows that LUCB-H matches the sample complexity of standard LUCB in misleading cases. It significantly reduces the sample complexity when offline data is reliable. Furthermore, we derived an instance-dependent lower bound that matches LUCB-H’s upper bound in certain cases. Experimental results demonstrated LUCB-H’s adaptability and robustness. It effectively reduces sample complexity in favorable cases while remaining robust when offline data is unreliable.

\paragraph{Acknowledgements} This research work is funded by the Singapore Ministry of Education AcRF Tier 2 grants (A-8000423-00-00 and A-8003063-00-00).

\bibliography{uai2025-template}

\begin{thebibliography}{29}
\providecommand{\natexlab}[1]{#1}
\providecommand{\url}[1]{\texttt{#1}}
\expandafter\ifx\csname urlstyle\endcsname\relax
  \providecommand{\doi}[1]{doi: #1}\else
  \providecommand{\doi}{doi: \begingroup \urlstyle{rm}\Url}\fi

\bibitem[Agrawal et~al.(2023)Agrawal, Juneja, Shanmugam, and Suggala]{agrawal2023optimal}
Shubhada Agrawal, Sandeep Juneja, Karthikeyan Shanmugam, and Arun~Sai Suggala.
\newblock Optimal best-arm identification in bandits with access to offline data.
\newblock \emph{arXiv preprint arXiv:2306.09048}, 2023.

\bibitem[Ben-David et~al.(2010)Ben-David, Blitzer, Crammer, Kulesza, Pereira, and Vaughan]{ben2010theory}
Shai Ben-David, John Blitzer, Koby Crammer, Alex Kulesza, Fernando Pereira, and Jennifer~Wortman Vaughan.
\newblock A theory of learning from different domains.
\newblock \emph{Machine learning}, 79:\penalty0 151--175, 2010.

\bibitem[Besbes et~al.(2022)Besbes, Ma, and Mouchtaki]{besbes2022beyond}
Omar Besbes, Will Ma, and Omar Mouchtaki.
\newblock Beyond iid: data-driven decision-making in heterogeneous environments.
\newblock \emph{Advances in Neural Information Processing Systems}, 35:\penalty0 23979--23991, 2022.

\bibitem[Blanchet et~al.(2019)Blanchet, Kang, and Murthy]{blanchet2019robust}
Jose Blanchet, Yang Kang, and Karthyek Murthy.
\newblock Robust wasserstein profile inference and applications to machine learning.
\newblock \emph{Journal of Applied Probability}, 56\penalty0 (3):\penalty0 830--857, 2019.

\bibitem[Bouneffouf et~al.(2019)Bouneffouf, Parthasarathy, Samulowitz, and Wistub]{bouneffouf2019optimal}
Djallel Bouneffouf, Srinivasan Parthasarathy, Horst Samulowitz, and Martin Wistub.
\newblock Optimal exploitation of clustering and history information in multi-armed bandit.
\newblock \emph{arXiv preprint arXiv:1906.03979}, 2019.

\bibitem[Bu et~al.(2020)Bu, Simchi-Levi, and Xu]{bu2020online}
Jinzhi Bu, David Simchi-Levi, and Yunzong Xu.
\newblock Online pricing with offline data: Phase transition and inverse square law.
\newblock In \emph{International Conference on Machine Learning}, pages 1202--1210. PMLR, 2020.

\bibitem[Chen et~al.(2022)Chen, Shi, and Pu]{chen2022data}
Xinyun Chen, Pengyi Shi, and Shanwen Pu.
\newblock Data-pooling reinforcement learning for personalized healthcare intervention.
\newblock \emph{arXiv preprint arXiv:2211.08998}, 2022.

\bibitem[Cheung and Lyu(2024)]{cheung2024leveraging}
Wang~Chi Cheung and Lixing Lyu.
\newblock Leveraging (biased) information: Multi-armed bandits with offline data.
\newblock \emph{arXiv preprint arXiv:2405.02594}, 2024.

\bibitem[Crammer et~al.(2008)Crammer, Kearns, and Wortman]{crammer2008learning}
Koby Crammer, Michael Kearns, and Jennifer Wortman.
\newblock Learning from multiple sources.
\newblock \emph{Journal of machine learning research}, 9\penalty0 (8), 2008.

\bibitem[Garivier and Kaufmann(2016)]{garivier2016optimal}
Aur{\'e}lien Garivier and Emilie Kaufmann.
\newblock Optimal best arm identification with fixed confidence.
\newblock In \emph{Conference on Learning Theory}, pages 998--1027. PMLR, 2016.

\bibitem[Gur and Momeni(2022)]{gur2022adaptive}
Yonatan Gur and Ahmadreza Momeni.
\newblock Adaptive sequential experiments with unknown information arrival processes.
\newblock \emph{Manufacturing \& Service Operations Management}, 24\penalty0 (5):\penalty0 2666--2684, 2022.

\bibitem[Hao et~al.(2023)Hao, Jain, Lattimore, Van~Roy, and Wen]{hao2023leveraging}
Botao Hao, Rahul Jain, Tor Lattimore, Benjamin Van~Roy, and Zheng Wen.
\newblock Leveraging demonstrations to improve online learning: Quality matters.
\newblock In \emph{International Conference on Machine Learning}, pages 12527--12545. PMLR, 2023.

\bibitem[Jamieson and Nowak(2014)]{jamieson2014best}
Kevin Jamieson and Robert Nowak.
\newblock Best-arm identification algorithms for multi-armed bandits in the fixed confidence setting.
\newblock In \emph{2014 48th annual conference on information sciences and systems (CISS)}, pages 1--6. IEEE, 2014.

\bibitem[Kalyanakrishnan et~al.(2012)Kalyanakrishnan, Tewari, Auer, and Stone]{kalyanakrishnan2012pac}
Shivaram Kalyanakrishnan, Ambuj Tewari, Peter Auer, and Peter Stone.
\newblock Pac subset selection in stochastic multi-armed bandits.
\newblock In \emph{ICML}, volume~12, pages 655--662, 2012.

\bibitem[Katz-Samuels and Scott(2019)]{katz2019top}
Julian Katz-Samuels and Clayton Scott.
\newblock Top feasible arm identification.
\newblock In \emph{The 22nd International Conference on Artificial Intelligence and Statistics}, pages 1593--1601. PMLR, 2019.

\bibitem[Kaufmann et~al.(2016)Kaufmann, Capp{\'e}, and Garivier]{kaufmann2016complexity}
Emilie Kaufmann, Olivier Capp{\'e}, and Aur{\'e}lien Garivier.
\newblock On the complexity of best-arm identification in multi-armed bandit models.
\newblock \emph{The Journal of Machine Learning Research}, 17\penalty0 (1):\penalty0 1--42, 2016.

\bibitem[Lai and Robbins(1985)]{lai1985asymptotically}
Tze~Leung Lai and Herbert Robbins.
\newblock Asymptotically efficient adaptive allocation rules.
\newblock \emph{Advances in applied mathematics}, 6\penalty0 (1):\penalty0 4--22, 1985.

\bibitem[Liu and Li(2016)]{liu2016prior}
Che-Yu Liu and Lihong Li.
\newblock On the prior sensitivity of thompson sampling.
\newblock In \emph{International Conference on Algorithmic Learning Theory}, pages 321--336. Springer, 2016.

\bibitem[Liu et~al.(2025)Liu, Dai, Zuo, Wang, Wong, Lui, and Chen]{liu2025offline}
Xutong Liu, Xiangxiang Dai, Jinhang Zuo, Siwei Wang, Carlee-Joe Wong, John Lui, and Wei Chen.
\newblock Offline learning for combinatorial multi-armed bandits.
\newblock \emph{arXiv preprint arXiv:2501.19300}, 2025.

\bibitem[Mansour et~al.(2009)Mansour, Mohri, and Rostamizadeh]{mansour2009domain}
Yishay Mansour, Mehryar Mohri, and Afshin Rostamizadeh.
\newblock Domain adaptation: Learning bounds and algorithms.
\newblock \emph{arXiv preprint arXiv:0902.3430}, 2009.

\bibitem[Russo and Van~Roy(2016)]{russo2016information}
Daniel Russo and Benjamin Van~Roy.
\newblock An information-theoretic analysis of thompson sampling.
\newblock \emph{Journal of Machine Learning Research}, 17\penalty0 (68):\penalty0 1--30, 2016.

\bibitem[Shivaswamy and Joachims(2012)]{shivaswamy2012multi}
Pannagadatta Shivaswamy and Thorsten Joachims.
\newblock Multi-armed bandit problems with history.
\newblock In \emph{Artificial Intelligence and Statistics}, pages 1046--1054. PMLR, 2012.

\bibitem[Si et~al.(2023)Si, Zhang, Zhou, and Blanchet]{si2023distributionally}
Nian Si, Fan Zhang, Zhengyuan Zhou, and Jose Blanchet.
\newblock Distributionally robust batch contextual bandits.
\newblock \emph{Management Science}, 69\penalty0 (10):\penalty0 5772--5793, 2023.

\bibitem[Simchowitz et~al.(2021)Simchowitz, Tosh, Krishnamurthy, Hsu, Lykouris, Dudik, and Schapire]{simchowitz2021bayesian}
Max Simchowitz, Christopher Tosh, Akshay Krishnamurthy, Daniel~J Hsu, Thodoris Lykouris, Miro Dudik, and Robert~E Schapire.
\newblock Bayesian decision-making under misspecified priors with applications to meta-learning.
\newblock \emph{Advances in Neural Information Processing Systems}, 34:\penalty0 26382--26394, 2021.

\bibitem[Wagenmaker and Pacchiano(2023)]{wagenmaker2023leveraging}
Andrew Wagenmaker and Aldo Pacchiano.
\newblock Leveraging offline data in online reinforcement learning.
\newblock In \emph{International Conference on Machine Learning}, pages 35300--35338. PMLR, 2023.

\bibitem[Wald(2004)]{wald2004sequential}
Abraham Wald.
\newblock \emph{Sequential analysis}.
\newblock Courier Corporation, 2004.

\bibitem[Wang et~al.(2021)Wang, Zhang, Singh, Riek, and Chaudhuri]{wang2021multitask}
Zhi Wang, Chicheng Zhang, Manish~Kumar Singh, Laurel Riek, and Kamalika Chaudhuri.
\newblock Multitask bandit learning through heterogeneous feedback aggregation.
\newblock In \emph{International Conference on Artificial Intelligence and Statistics}, pages 1531--1539. PMLR, 2021.

\bibitem[Ye et~al.(2020)Ye, Lin, Xie, and Lui]{ye2020combining}
Li~Ye, Yishi Lin, Hong Xie, and John Lui.
\newblock Combining offline causal inference and online bandit learning for data driven decision.
\newblock \emph{arXiv preprint arXiv:2001.05699}, 2020.

\bibitem[Zhang et~al.(2019)Zhang, Agarwal, Daum{\'e}~III, Langford, and Negahban]{zhang2019warm}
Chicheng Zhang, Alekh Agarwal, Hal Daum{\'e}~III, John Langford, and Sahand~N Negahban.
\newblock Warm-starting contextual bandits: Robustly combining supervised and bandit feedback.
\newblock \emph{arXiv preprint arXiv:1901.00301}, 2019.

\end{thebibliography}

\newpage

\onecolumn

\title{Best Arm Identification with Biased Offline Data\\(Supplementary Material)}
\maketitle

\appendix
\section{Preliminaries}
\begin{proposition}
    (Chernoff Inequality). Let \( G_1, \dots, G_m \) be independent (though not necessarily identically distributed) 1-subGaussian random variables. For any \( \delta \in (0,1) \), it holds that
    \begin{align*}
        \Pr \left[ \left|\frac{1}{m} \mathbb{E} \left[ \sum_{i=1}^{m} G_i \right] - \frac{1}{m} \sum_{i=1}^{m} G_i\right|\leq \sqrt{\frac{2 \log(k/\delta)}{m}} \right] \geq 1 - \frac{\delta}{k},
    \end{align*}
\end{proposition}
We consider an identifiable class of bandit models of the form:
\begin{equation}\label{instances}
    \mathcal{M}=\{v=(v_{1},v_{2},\ldots,v_{k}):v_{i}\in \mathcal{P},\mu^{\text{on}}(1)>\mu^{\text{on}}(2)\geq \ldots\geq \mu^{\text{on}}(k)\}
\end{equation}
For all instances in set $\mathcal{M}$, $I^{*}=1$. Let $\mathcal{A}=((\pi_{t}),\tau,\hat{I}^{*})$ be a $\delta$-PAC algorithm. 

Let $I_P = (P^{\mathrm{off}}, P^{\mathrm{on}})$ and $I_Q = (Q^{\mathrm{off}}, Q^{\mathrm{on}})$ be two problem instances defined over a common finite arm set $A$. Both instances share the same offline sample sizes $\{T_S(i)\}_{i\in A}$ and confidence level $\delta$, but may differ in their offline and online reward distributions. 
Let $\tau_{\delta}(Q, i)$ be the (random) number of online samples drawn from arm $i$ under instance $I_Q$ before the stopping time $\tau_\delta$; we write $\mathbb{E}[\tau_{\delta}(Q, i)]$ for its expectation. The probability that an event $\mathcal{E}$ occurs under instance $I_Q$ and instance $I_P$ at stopping time $\sigma$ are written as $\mathbb{P}_{\sigma}(Q, \mathcal{E})$ and $\mathbb{P}_{\sigma}(P, \mathcal{E})$ respectively. The Kullback-Leibler (KL) divergence between distributions $P$ and $Q$ is written as $\mathrm{KL}(P, Q)$, and $d(p, q)$ denotes the KL divergence measure between Bernoulli distributions with means $p$ and $q$. We assume that the stopping time $\sigma$ is adapted to the filtration $\{\mathcal{F}_t\}_{t \geq 0}$ and is almost surely finite. To establish the theoretical results, we first introduce the following lemma.

\begin{lemma}\label{ChangeofDistribution}
Let $\mathcal{E} \in \mathcal{F}_\sigma$ be any event measurable with respect to the stopping time $\sigma$. For any non-anticipatory policy $\pi$, the following inequality holds:
\begin{align}
\sum_{i=1}^{k} \mathbb{E}[N_{\sigma}(i)] \, \mathrm{KL}\bigl(Q^{\mathrm{on}}(i), P^{\mathrm{on}}(i)\bigr) + \sum_{i=1}^{k} T_{S}(i)\, \mathrm{KL}\bigl(Q^{\mathrm{off}}(i), P^{\mathrm{off}}(i)\bigr) \geq d\bigl(\mathbb{P}_{\sigma}(Q, \mathcal{E}), \mathbb{P}_{\sigma}(P, \mathcal{E})\bigr).
\end{align}
\end{lemma}

\emph{Proof}:
The log-likelihood ratio for the sequence of observations observed by time $t$ under an algorithm $\mathcal{A}$ is
\begin{align*}
    L_{t}=L(\{X_{s}(i)\}_{s=1}^{T_{S}(i)}, A_{1},\ldots,A_{t}, Y_{A_{1}}, \ldots,Y_{A_{t}})\coloneqq\sum_{i=1}^{k}\sum_{l=1}^{t}\mathbbm{1}\{A_{l}=i\}\log\left(\frac{Q^{\mathrm{on}}_{i}(Y_{A_{l}})}{P^{\mathrm{on}}_{i}(Y_{A_{l}})}\right)+\sum_{i=1}^{k}\sum_{s=1}^{T_{S}(i)}\log\left(\frac{Q^{\mathrm{off}}_{i}(X_{s})}{P^{\mathrm{off}}_{i}(X_s)}\right).
\end{align*}
Recall that $\{X_{s}(i)\}_{s=1}^{T_{S}(i)}$ are i.i.d.\ samples from the offline distribution $P^{\mathrm{off}}_{i}$, while $\{Y_{l}(i)\}_{l=1}^{N_{\sigma}(i)}$ are i.i.d. samples from the online distribution $P^{\mathrm{on}}_{i}$. We can rewritten $L_{\sigma}$ as
\begin{align*}
    L_{\sigma}=\sum_{i=1}^{k}\sum_{l=1}^{N_{\sigma}(i)}\log\left(\frac{Q^{\mathrm{on}}_{i}(Y_{A_{l}})}{P^{\mathrm{on}}_{i}(Y_{A_{l}})}\right)+\sum_{i=1}^{k}\sum_{s=1}^{T_{S}(i)}\log\left(\frac{Q^{\mathrm{off}}_{i}(X_{s})}{P^{\mathrm{off}}_{i}(X_s)}\right).
\end{align*}
Since $\mathbb{E}_{Q}\left[\log\left(\frac{Q^{\mathrm{on}}_{i}(Y_{A_{l}})}{P^{\mathrm{on}}_{i}(Y_{A_{l}})}\right)\right]=\mathrm{KL}(Q^{\mathrm{on}}(i), P^{\mathrm{on}}(i))$ and $\mathbb{E}_{Q}\left[\log\left(\frac{Q^{\mathrm{off}}_{i}(X_{s})}{P^{\mathrm{off}}_{i}(X_s)}\right)\right]=\mathrm{KL}(Q^{\mathrm{off}}(i), P^{\mathrm{off}}(i))$, by Wald's Lemma \citep{wald2004sequential},
\begin{align}\label{ex}
    \mathbb{E}_{Q}[L_{\sigma}]=\sum_{i=1}^{k} \mathbb{E}[N_{\sigma}(i)] \, \mathrm{KL}\bigl(Q^{\mathrm{on}}(i), P^{\mathrm{on}}(i)\bigr) + \sum_{i=1}^{k} T_{S}(i)\, \mathrm{KL}\bigl(Q^{\mathrm{off}}(i), P^{\mathrm{off}}(i)\bigr).
\end{align}
By Lemma 19 in \citet{kaufmann2016complexity}, we have
\begin{align}\label{19}
    \mathbb{E}_{Q}[L_{\sigma}]\geq d\bigl(\mathbb{P}_{\sigma}(Q, \mathcal{E}), \mathbb{P}_{\sigma}(P, \mathcal{E})\bigr).
\end{align}
Combining (\ref{ex}) and (\ref{19}) completes the proof.


\section{Proof of Proposition \ref{impossible} } \label{app:proof_prop}
\emph{Proof}:
In the instance $I_{P}$, arm $1$ is the best arm, and in the instance $I_{Q}$, arm $2$ is the best arm. In both instances $I_P, I_Q $, the DM requires $\mathbb{E}[\tau_{\delta}']= \Omega(\log (1/\delta) / \delta^{2\beta})$ pulls (without offline data) to be $\delta$-PAC. Let $\sigma$ be any almost surely finite stopping time with respect to $\mathcal{F}_{t}$ on instance $I_{Q}$. Introducing the event $\mathcal{E}=(\hat{I}^{*}=2)\in \mathcal{F}_{\sigma}$, any $\delta$-PAC algorithm satisfies $\mathbb{P}_{\sigma}(Q,\mathcal{E})\geq 1-\delta$ and $\mathbb{P}_{\sigma}(P,\mathcal{E})<\delta/2$, where $\mathbb{P}_{\sigma}(P,\mathcal{E})$ and $\mathbb{P}_{\sigma}(Q,\mathcal{E})$ are the probabilities of the event $\mathcal{E}$ occurring on instances $I_{P}$ and $I_{Q}$ separately in round $\sigma$. Denote $\mathbb{E}_{\sigma}(Q,i)$ as the expectation number of samples allocated to arm $i$ until round $\sigma$ on instance $I_{Q}$, where $i\in \{1,2\}$.

By Lemma \ref{ChangeofDistribution}, we have
\begin{align*}
    d(\mathbb{P}_{\sigma}(Q,\mathcal{E}),\mathbb{P}_{\sigma}(P,\mathcal{E}))\leq &\sum_{i=1}^{2}\mathbb{E}_{\sigma}(Q,i)\mathrm{KL}(Q^{\mathrm{on}}(i),P^{\mathrm{on}}(i))
    +\sum_{i=1}^{2}T_{S}(i)\mathrm{KL}(Q^{\mathrm{off}}(i),P^{\mathrm{off}}(i)),
\end{align*}
where the function $d(x, y) := x \log\left(x/y\right) + (1 - x) \log\left((1-x)/(1-y)\right)$ is the binary relative entropy, with the convention that \(d(0, 0) = d(1, 1) = 0\).
Then the following inequality holds, 
\begin{align*}
    \mathbb{E}_{\sigma}(Q,2)&\geq \frac{d(1-\delta,\frac{\delta}{2})-T_{S}(2)\mathrm{KL}(Q^{\mathrm{off}}(2),P^{\mathrm{off}}(2))}{\mathrm{KL}(Q^{\mathrm{on}}(2),P^{\mathrm{on}}(2))}
\end{align*}
due to the fact that $ Q_{1}^{\mathrm{off}}=P_{1}^{\mathrm{off}}$ and $Q_{1}^{\mathrm{on}}=P_{1}^{\mathrm{on}}$. 
The monotonicity properties of $d(x,y)$ is increasing when $x>y$ and decreasing when $x<y$ yield $d(1-\delta,\delta/2)>d(1-\delta,\delta)$. Hence, we can find an $\epsilon'>0$ such that $d(1-\delta,\delta/2)>\delta^{-\epsilon'} d(1-\delta,\delta)$. Then
\begin{align*}
    &\mathbb{E}_{\sigma}(Q,2)
    \geq 2\delta^{-2\beta-\epsilon'}\log\Big(\frac{1}{2.4\delta}\Big)-\frac{T_{S}(2)\mathrm{KL}(Q^{\mathrm{off}}(2),P^{\mathrm{off}}(2))}{\mathrm{KL}(Q^{\mathrm{on}}(2),P^{\mathrm{on}}(2))}\\
    &\quad\geq 2\delta^{-2\beta-\epsilon'}\log\Big(\frac{1}{2.4\delta}\Big)-\delta^{-2\beta-\epsilon'}\log\Big(\frac{1}{\delta}\Big)
    =\Omega\bigg(\delta^{-2\beta-\epsilon'}\log\Big(\frac{1}{\delta}\Big)\bigg).
\end{align*}
The second inequality holds by the claim assumption that we can find $\epsilon''>0$ such that $T_{S}(2)\leq \frac{C}{2}(\delta^{-2\beta}-\delta^{-2\beta+\epsilon''})$. Let $\epsilon=\{\epsilon', \epsilon'', \beta\}$. Hence we have
\begin{align*}
    \mathbb{E}[\tau_{\delta}(Q)]\geq \Omega\bigg(\delta^{-2\beta-\epsilon}\log\Big(\frac{1}{\delta}\Big)\bigg)
\end{align*}
as desired. 

\section{Proof of Theorem \ref{thm: 4.1}} \label{app:proof_ach}
To prove Theorem \ref{thm: 4.1}, we  first state a lemma.
\begin{lemma}\label{confidence}
    The following events hold over all rounds $t=1,2,\ldots$ with probability at least $1-\delta/k$ :
    $$\bigg|\min(\mathrm{UCB}_t^S(i), \mathrm{UCB}_{t}(i))-\mu^{\mathrm{on}}(i)\bigg|\leq  \min\bigg(2\sqrt{\frac{2 \log(kt /\delta)}{N_t(i) + T_S(i)}} + \frac{T_S(i)}{N_t(i) + T_S(i)}  \cdot \eta(i) , 2\sqrt{\frac{2 \log(kt /\delta)}{N_t(i)}}\bigg)$$ 
    and \\
    $$\bigg|\mu^{\mathrm{on}}(i)-\max(\mathrm{LCB}_t^S(i), \mathrm{LCB}_{t}(i))\bigg|\leq  \min\bigg(2\sqrt{\frac{2 \log(kt /\delta)}{N_t(i) + T_S(i)}} + \frac{T_S(i)}{N_t(i) + T_S(i)}  \cdot \eta(i) , 2\sqrt{\frac{2 \log(kt /\delta)}{N_t(i)}}\bigg).$$
\end{lemma}

\emph{Proof}:
To prove this Lemma, it suffices to prove that 
\begin{align*}
    \mathbb{P}\left\{ \mu^{(\mathrm{on})} (i) \leq \mathrm{UCB}_t^S (i) \leq \mu^{(\mathrm{on})} (i) + 2\sqrt{\frac{2 \log(k t/\delta)}{N_t(i) + T_S(i)}} + \frac{T_S(i)}{N_t(i) + T_S(i)}  \cdot \eta(i)  \right\}\geq 1-\frac{\delta}{kt}.
\end{align*}
Indeed, the inequality implies 
\begin{align*}
    \mathbb{P}\left\{ \mu^{(\mathrm{on})} (i) \leq \mathrm{UCB}_t^S (i) \leq \mu^{(\mathrm{on})} (i) + 2\sqrt{\frac{2 \log(kt /\delta)}{N_t(i) }}  \right\}\geq 1-\frac{\delta}{kt}
\end{align*}
by setting $T_{S}(i)=0$.
\begin{align*}
    &\mathbb{P}[\mu^{(\mathrm{on})}(i) \leq \mathrm{UCB}_t^S(i)] \\
    &= \mathbb{P} \left[ \mu^{(\mathrm{on})}(i) \leq \frac{N_t(i) \cdot \hat{Y}_t(i) + T_S(i) \cdot \hat{X}(i)}{N_t(i) + T_S(i)} + \sqrt{\frac{2\log(kt /\delta)}{N_t^+(i) + T_S(i)}} + \frac{T_S(i)}{N_t^+(i) + T_S(i)} \cdot V(i) \right] \\
    &= \mathbb{P} \left[ \frac{N_t(i) \mu^{(\mathrm{on})}(i) + T_S(i) \mu^{(\mathrm{off})}(i)}{N_t(i) + T_S(i)} 
   \frac{T_S(i)(\mu^{(\mathrm{on})}(i) - \mu^{(\mathrm{off})}(i))}{N_t(i) + T_S(i)} 
    \leq \frac{N_t(i) \cdot \hat{Y}_t(i) + T_S(i) \cdot \hat{X}(i)}{N_t(i) + T_S(i)} \right.\\
    &\quad\left. + \sqrt{\frac{2\log(kt /\delta)}{N_t(i) + T_S(i)}} + \frac{T_S(i)}{N_t(i) + T_S(i)} \cdot V(i) \right] \\
    &\geq  \mathbb{P} \left[ \frac{N_t(i) \mu^{(\mathrm{on})}(i) + T_S(i) \mu^{(\mathrm{off})}(i)}{N_t(i) + T_S(i)} 
     \leq \frac{N_t(i) \cdot \hat{Y}_t(i) + T_S(i) \cdot \hat{X}(i)}{N_t(i) + T_S(i)} 
    + \sqrt{\frac{2\log(kt /\delta)}{N_t(i) + T_S(i)}} \right] \\
    &\geq \mathbb{P}_{Y_i \sim P^{(\mathrm{on})}(i)} \left[ \frac{n \mu^{(\mathrm{on})}(i) + T_S(i) \mu^{(\mathrm{off})}(i)}{n + T_S(i)} 
   \leq \frac{\sum_{i=1}^{n} Y_i + T_S(i) \cdot \hat{X}(i)}{n + T_S(i)} 
     + \sqrt{\frac{2\log(kt /\delta)}{n + T_S(i)}} \quad \mathrm{for }\quad n = 1,2,\dots,t \right] \\
    &\geq 1 -\frac{\delta}{k}
\end{align*}
and
\begin{align*}
    &\mathbb{P} \Bigg( \mathrm{UCB}_t^S(i) \leq \mu^{(\mathrm{on})}(i) + 2\sqrt{\frac{2 \log(kt /\delta)}{N_t(i) + T_S(i)}} + \frac{T_S(i)}{N_t(i) + T_S(i)}  \cdot \eta(i)  \Bigg) \\
    &= \mathbb{P} \Bigg( \frac{N_t(i) \cdot \hat{Y}_t(i) + T_S(i) \cdot \hat{X}(i)}{N_t(i) + T_S(i)} \leq \mu^{(\mathrm{on})}(i)  + \sqrt{\frac{2\log(kt/\delta)}{N_t^+(i) + T_S(i)}} + \frac{T_S(i) \cdot (\mu^{(\mathrm{off})}(i) - \mu^{(\mathrm{on})}(i))}{N_t^+(i) + T_S(i)} \Bigg) \\
    &= \mathbb{P} \Bigg( \frac{N_t(i) \cdot \hat{Y}_t(i) + T_S(i) \cdot \hat{X}(i)}{N_t(i) + T_S(i)} \leq \frac{N_t(i) \mu^{(\mathrm{on})}(i) + T_S(i) \mu^{(\mathrm{off})}(i)}{N_t(i) + T_S(i)} 
    + \sqrt{\frac{2\log(kt/\delta)}{N_t^+(i) + T_S(i)}} \Bigg) \\
    &\geq \mathbb{P}_{Y_i \sim P^{(\mathrm{on})}(i)} \Bigg[ \frac{\sum_{i=1}^{n} Y_i + T_S(i) \cdot \hat{X}(i)}{n + T_S(i)} \leq \frac{n \mu^{(\mathrm{on})}(i) + T_S(i) \mu^{(\mathrm{off})}(i)}{n + T_S(i)}  + \sqrt{\frac{2\log(kt/\delta)}{n + T_S(i)}} \quad \mathrm{for } \quad n = 1,2,\dots,t \Bigg] \\
    &\geq 1 -\frac{\delta}{k}.
\end{align*}
Then a union bound on the failure probabilities for $i\in A$ establishes the Lemma. We can obtain a similar bound for $\big|\mu^{\mathrm{on}}(i)-\max(\mathrm{LCB}_t^S(i), \mathrm{LCB}_{t}(i))\big|$. Clearly, the stopping condition guarantees that the probability of identifying the best arm is no less than $1 - \delta$.

In the following, we bound the total number of measurements.
Define $c=\frac{1}{2}[ \mu^{\mathrm{on}}(1)+\mu^{\mathrm{on}}(2)]$. We say that arm $i^*$ is BAD in round $t$, if $\mathrm{LCB}_{t}(i^*)<c$ and $\mathrm{LCB}_{t}^{S}(i^*)<c$, and an arm $i\neq i^*$ is BAD in round $t$, if $\mathrm{UCB}_{t}(i)>c$ and $\mathrm{UCB}_{t}^{S}(i)>c$. We claim that for all time $t\geq 1$, 
\begin{align*}
    \big\{\text{Events in Lemma } \ref{confidence} \text{ hold }\big\} \cap \big\{\max\{\mathrm{LCB}_{t}(i^*), \mathrm{LCB}_{t}^{S}(i^*)\}<\min\{\mathrm{UCB}_{t}(i), \mathrm{UCB}_{t}^{S}(i)\}\big\}\\
    \Rightarrow \{h_{t} \text{ is BAD }\} \cap \{l_{t} \text{ is BAD }\}.
\end{align*}

Assume that the events in  Lemma \ref{confidence} hold, then for any $i\neq i^*$ and $s\geq \tau_{i}$, where $\tau_{i}$ is the first integer such that arm $i$ is not BAD. Define $\tau_{i^*}=\tau_{2}$.
\begin{align*}
        \mathrm{UCB}_{t}(i)&\leq \mu^{(\mathrm{on})}(i) + 2\ \mathrm{rad}_t(i)  \\
        &= c + 2\ \mathrm{rad}_t(i) + \frac{(\mu^{\mathrm{on}}(i)-\mu^{\mathrm{on}}(i^*))+(\mu^{\mathrm{on}}(i)-\mu^{\mathrm{on}}(2))}{2}\\
        &\leq c + 2\ \mathrm{rad}_t(i) - \frac{\Delta_{i}}{2}\leq c, 
\end{align*}
where $\Delta_{i}=\mu^{\mathrm{on}}(i^*)-\mu^{\mathrm{on}}(i)$.
\begin{align*}
         \mathrm{UCB}_t^S(i) &\leq \mu^{(\mathrm{on})}(i) + \mathrm{rad}_t^S(i) + \left[ \sqrt{\frac{2 \log(kt / \delta)}{N_t(i) + T_S(i)}} + \frac{T_S(i) \cdot \left(\mu^{(\mathrm{off})}(i) - \mu^{(\mathrm{on})}(i)\right)}{N_t(i) + T_S(i)} \right]\\
       & =c -\frac{\Delta_{i}}{2} + 2\sqrt{\frac{2 \log(kt / \delta)}{N_t(i) + T_S(i)}} + \frac{T_S(i)}{N_t(i) + T_S(i)}  \cdot \eta(i) \leq c. 
\end{align*}
Since $\min \{ \mathrm{UCB}_{s}(i),\mathrm{UCB}_t^S(i) \} \leq c$, we have
\begin{align*}
    \min\bigg\{2\sqrt{\frac{2 \log(k t/\delta)}{N_t(i) + T_S(i)}} + \frac{T_S(i)}{N_t(i) + T_S(i)}  \cdot \eta(i) , 2\sqrt{\frac{2 \log(kt /\delta)}{N_t(i)}}\bigg\}\leq \frac{\Delta_{i}}{2}.
\end{align*}
Solving for $N_{t}(i)$, we find that
\begin{align*}
    N_{t}(i)>\frac{32}{\Delta_{i}^{2}}\log\Big(\frac{k}{\delta}\Big)-T_{S}(i)\cdot\max\Big\{1-\frac{4\eta(i)}{\Delta_{i}},0\Big\}.
\end{align*}
Since Lemma \ref{confidence} holds, the total number of rounds is observed to be no greater than
\begin{align*}
\sum_{t=1}^{\infty} \mathbbm{1}\{ h_t \text{ is } \mathrm{BAD} \text{ or } l_t \text{ is } \mathrm{BAD} \}
&= \sum_{t=1}^{\infty} \sum_{i=1}^{n} \mathbbm{1}\left\{ \{ h_t = i \text{ or } l_t = i \} \cap \{ i \text{ is } \textit{BAD} \} \right\} \\
&\leq \sum_{t=1}^{\infty} \sum_{i=1}^{n} \mathbbm{1}\left\{ \{ h_t = i \text{ or } l_t = i \} \cap \{ N_t(i) \leq \tau_i \} \right\} 
\leq \sum_{i=1}^{n} \tau_i
\end{align*}
Then LUCB-H algorithm obtains a sample complexity of order in Theorem $\ref{thm: 4.1}$.
\section{Proof of Theorem \ref{thm 5.1}}\label{app:proof_conv}
Fix $a>0$. For all $i\in\{1,2,\ldots,k\}$, from Assumption \ref{continuity} there exists an alternative model
\begin{align*}
    v'=(v_{1},\ldots,v_{i-1},v_{i}',v_{i+1},\ldots,v_{k})
\end{align*}
in which the only arm modified is arm $i$, $i\neq 1$ and $v_{i}'$ is such that:
\begin{align*}
    \mathrm{KL}(v_{i},v_{1})<\mathrm{KL}(v_{i},v_{i}')<\mathrm{KL}(v_{i},v_{1})+a, \quad\mu^{\mathrm{on'}}(i)>\mu^{\mathrm{on'}}(1)
\end{align*}
and 
\begin{align*}
\mu^{\mathrm{off'}}(i) = 
\begin{cases}
\mu^{\mathrm{off}}(i), & \text{if } \mu^{\mathrm{off}}(i) > \mu^{\mathrm{off}}(1) \\
\mu^{\mathrm{off}}(1), & \text{if } \mu^{\mathrm{off}}(i) < \mu^{\mathrm{off}}(1).
\end{cases}
\end{align*}
In particular, on the bandit model $v'$ the best arm is no longer arm $1$. Introducing the event $\mathcal{E}=\{\hat{I}^{*}=1 \}\in \mathcal{F}_{\tau_{\delta}}$, any $\delta$-PAC algorithm satisfies $\mathbb{P}_{\tau_{\delta}}(v,\mathcal{E})\geq 1-\delta$ and $\mathbb{P}_{\tau_{\delta}}(v',\mathcal{E})\le\delta$.
By Lemma \ref{ChangeofDistribution}, we have 
\begin{align*}
        \sum_{i=1}^{k}\mathbb{E}_{v}[N_{\tau_{\delta}}(i)]\mathrm{KL}(v^{\mathrm{on}}(i),v^{\mathrm{on}'}(i))+\sum_{i=1}^{k}T_{S}(i)\mathrm{KL}(v^{\mathrm{off}}(i),v^{\mathrm{off}'}(i))\geq d\bigl(\mathbb{P}_{\tau_{\delta}}(v,\mathcal{E}), \mathbb{P}_{\tau_{\delta}}(v',\mathcal{E})\bigr)\geq d(1-\delta,\delta).
\end{align*}
Then the following holds
\begin{align*}
    \sum_{i=1}^{k}\mathbb{E}_{v}[N_{\tau_{\delta}}(i)]\mathrm{KL}(v^{\mathrm{on}}(i),v^{\mathrm{on}'}(i))+\sum_{i=1}^{k}T_{S}(i)\mathrm{KL}(v^{\mathrm{off}}(i),v^{\mathrm{off}'}(i))\geq \log\Big(\frac{1}{2.4\delta}\Big),
\end{align*}
due to  the fact that $d(1-\delta,\delta)\geq\log(1/(2.4\delta))$. 
Thus
\begin{align*}
     \mathbb{E}_{v}[N_{\tau_{\delta}}(i)]&\geq \frac{\log(1/(2.4\delta))-T_{S}(i)\mathrm{KL}(v^{\mathrm{off}}(i),v^{\mathrm{off}'}(i))}{\mathrm{KL}(v^{\mathrm{on}}(i),v^{\mathrm{on}'}(i))+a}\\
    &=\frac{\log(1/(2.4\delta))}{\mathrm{KL}(v^{\mathrm{on}}(i),v^{\mathrm{on}'}(i))+a}-T_{S}(i)\frac{\mathrm{KL}(v^{\mathrm{off}}(i),v^{\mathrm{off}'}(i))}{\mathrm{KL}(v^{\mathrm{on}}(i),v^{\mathrm{on}'}(i))+a}\\
    &=\frac{\log(1/(2.4\delta))}{\mathrm{KL}(v^{\mathrm{on}}(i),v^{\mathrm{on}'}(i))+a}-T_{S}(i)\cdot\max\bigg\{\frac{\mu^{\mathrm{off}}(1)-\mu^{\mathrm{off}}(i)}{\Delta_{i}},0\bigg\}^{2}
\end{align*}
The last equality holds  because $\mathrm{KL}(v^{\mathrm{off}}(i),v^{\mathrm{off}'}(i))=\max\{0, \mu^{\mathrm{off}}(1)-\mu^{\mathrm{off}}(i)\}^{2}/2$.
Letting $a$ tend to zero and summing over the arms yields the bound on $\mathbb{E}_{v}[\tau_\delta]=\sum_{i=1}^{k}\mathbb{E}_{v}[N_{\tau_{\delta}}(i)]$.

\begin{figure*}[t]
    \centering
    \begin{minipage}{0.42\textwidth}  
        \centering
        \includegraphics[width=\textwidth]{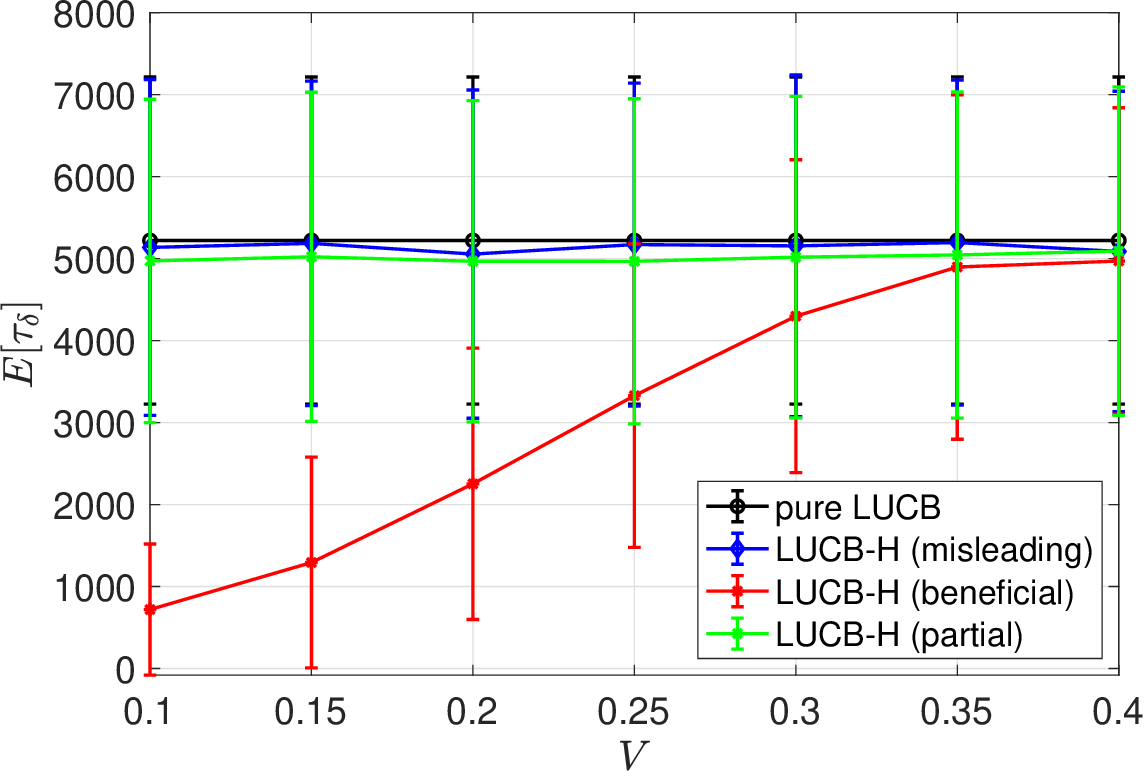}
        \caption{Evolution of $\mathbb{E}[\tau_{\delta}]$ with $V$ in group 2.}
        \label{fig:fig9}
    \end{minipage}
    \hspace{0.04\textwidth}  
    \begin{minipage}{0.42\textwidth}  
        \centering
        \includegraphics[width=\textwidth]{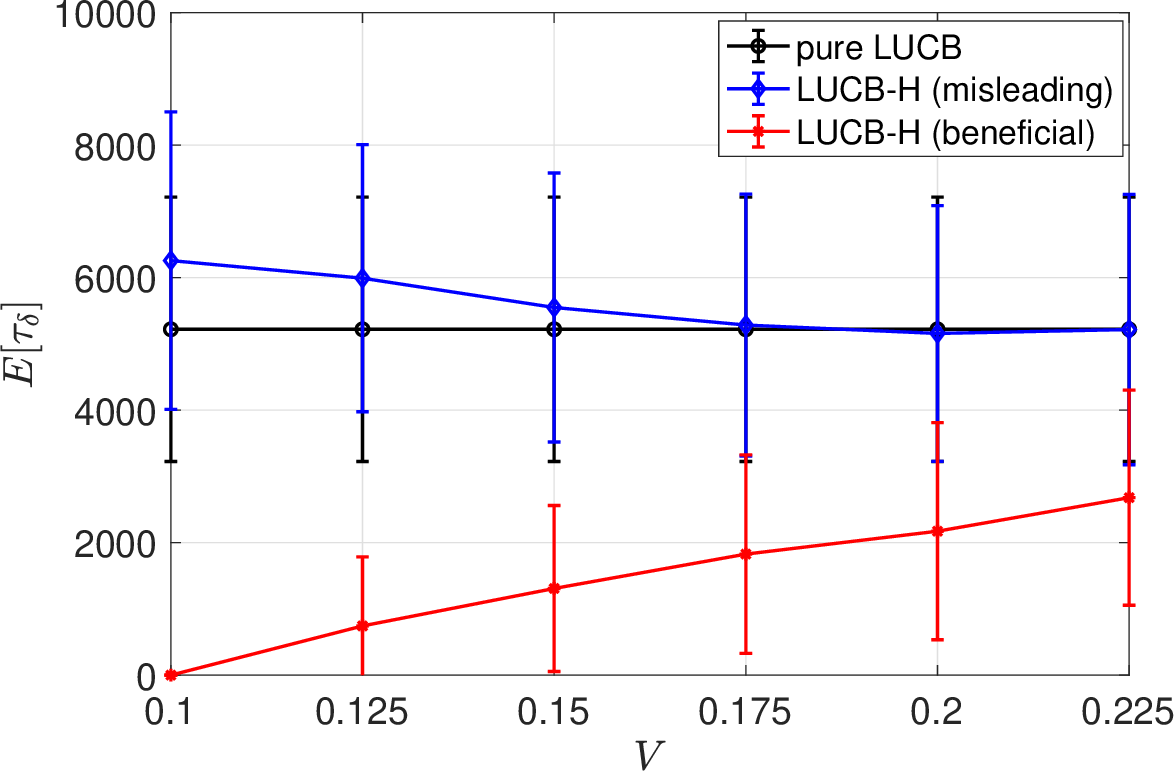}
        \caption{Evolution of $\mathbb{E}[\tau_{\delta}]$ with $V$ in group 2.}
        \label{fig:fig10}
    \end{minipage}
\end{figure*}

\section{Robustness to Misspecifications of the Bias Bound $V(i)$}  \label{app:robustnes}
The LUCB-H algorithm relies on prior knowledge of an auxiliary bias bound $V(i)$, which serves as an upper bound on the mean shift between the offline and online reward distributions. That is, $V(i) \geq |\mu^{\mathrm{on}}(i) - \mu^{\mathrm{off}}(i)|$ for all arms $i$. In many real-world applications, it is possible to approximate or conservatively upper bound $V(i)$ before the online phase begins. For instance, in recommendation systems or clinical trials, offline and online data often originate from similar but not identical distributions. Practitioners may exploit prior logs, domain expertise, or distribution shift estimation techniques to estimate such discrepancies. We have provided a brief discussion of potential empirical strategies in the last paragraph of Page 2.

Section 6 has already demonstrated the strong empirical performance of LUCB-H under correctly specified bias bounds. In this section, we investigate how LUCB-H performs when $V(i)$ is misestimated, particularly when it is underestimated. This corresponds to the case where $V(i) < |\mu^{\mathrm{on}}(i) - \mu^{\mathrm{off}}(i)|$, which could potentially compromise the algorithm’s ability to properly discount misleading offline data.

To assess the sensitivity of LUCB-H to misestimation of $V(i)$, we conducted an additional experiment in Group 2 (see Figure \ref{fig:fig9}), where we set $V(i) \in \{0.1, 0.15\}$ for suboptimal arms while the true discrepancy was $0.2$. The results show that even with $V(i)$ set below the true bias, LUCB-H remains stable and competitive. This suggests that LUCB-H is empirically robust to mild underestimation of the bias.

To better understand when such underestimation can lead to performance degradation, we provide the following analysis. In the extreme case where $V(i) = 0$, LUCB-H fully trusts offline data and assumes perfect alignment between offline and online distributions. Consequently, the selection rule $\min\{\mathrm{UCB}_t(i), \mathrm{UCB}^S_t(i)\}$ may favor the offline-based bound $\mathrm{UCB}^S_t(i)$, causing the algorithm to overly rely on historical samples. This behavior is particularly problematic if arm $i$ is the true best arm and the offline mean is significantly biased, leading to suboptimal performance. On the other hand, if arm $i$ is suboptimal, relying on biased offline estimates may reduce its chances of being pulled, which could actually help reduce sample complexity.

In Group 2's previous experiments, we fixed $V(1) = 0.4$ for the best arm and only varied $V(i)$ for suboptimal arms. Since $V(1)$ was not underestimated, LUCB-H maintained its effectiveness. To further investigate the impact of bias misestimation on the best arm, we conducted a follow-up experiment in which all arms share the same bias bound, varying $V(i) \in \{0.1, 0.125, 0.15, 0.175, 0.2, 0.225\}$ for $i=1,2,\ldots,5$ (see Figure \ref{fig:fig10}). This setup allows us to explore both underestimation and overestimation scenarios.

The results confirm that LUCB-H is robust when $V(i)$ is overestimated: its performance remains comparable to that of Pure  LUCB. However, when $V(i)$ is underestimated, especially for the best arm, LUCB-H may be misled by biased offline data and underperform relative to the pure online baseline LUCB. These findings provide empirical support for the theoretical necessity of a valid bias bound, as emphasized in our impossibility result in Proposition~\ref{impossible}. Without reliable information to distinguish helpful from misleading historical data, no algorithm can consistently outperform the conservative online LUCB baseline across all problem instances. A valid $V(i)$ enables LUCB-H to safely incorporate historical data when appropriate and revert to cautious online behavior when necessary.

\end{document}